%% file: root.tex
\documentclass[11pt]{eptcs}
\pdfoutput=1

\usepackage{amsmath} 
\usepackage{amssymb}  
\usepackage{siunitx}
\usepackage{tikz}
\usepackage{wrapfig}
\usepackage{algpseudocode,algorithm,algorithmicx}
\usepackage{tabularx}
\usepackage{dblfloatfix}

\usepackage{outlines}

\usepackage{float}
\usepackage{color}  
\usepackage{xspace}
\usepackage{tabularx}
\usepackage{multirow}
\usepackage{listings}
\usepackage{amsfonts,amsmath}
\usepackage{tabulary}
\newcolumntype{K}[1]{>{\centering\arraybackslash}p{#1}}
\usepackage{enumitem}
\setenumerate[1]{label=\Roman*.}
\setenumerate[2]{label=\Alph*.}
\setenumerate[3]{label=\roman*.}
\setenumerate[4]{label=\alph*.}
\usepackage{xspace}
\usepackage{siunitx}
\usepackage{tikz}
\usetikzlibrary{arrows,calc,decorations.pathmorphing,decorations.markings,backgrounds,decorations,decorations.pathmorphing,positioning,fit,automata,shapes,patterns,decorations.text}
\usepackage{tabularx}
\usepackage{rotating}
\usepackage{outlines}
\usepackage{enumitem}
\usepackage{graphicx} 
\usepackage{color}  
\usepackage{adjustbox}
\usepackage{array}
\usepackage{algpseudocode,algorithm,algorithmicx}
\usepackage{tuncali-dissertation-macros}

\providecommand{\publicationstatus}{This article is a modified version of a chapter from Cumhur Erkan Tuncali's Ph.D. Dissertation \cite{tuncali2019dissertation}.}

\title{\LARGE \bf
Rapidly-exploring Random Trees-based Test Generation for Autonomous Vehicles
\thanks{This work was partially funded by NSF award CNS 1350420}
}

\author{Cumhur Erkan Tuncali \quad\ Georgios Fainekos
		\institute{Arizona State University, Tempe, AZ, USA}
	\email{etuncali, fainekos@asu.edu}}
	
\def\titlerunning{RRT-based Test Generation for Autonomous Vehicles}
\def\authorrunning{C.E.~Tuncali, G.~Fainekos}

\begin{document}

\maketitle

\begin{abstract}
Autonomous vehicles are in an intensive research and development stage, and the organizations developing these systems are targeting to deploy them on public roads in a very near future.
One of the expectations from fully-automated vehicles is never to cause an accident.
However, an automated vehicle may not be able to avoid all collisions, \eg the collisions caused by other road occupants.
Hence, it is important for the system designers to understand the boundary case scenarios where an autonomous vehicle can no longer avoid a collision.
In this paper, an automated test generation approach that utilizes Rapidly-exploring Random Trees is presented.
A comparison of the proposed approach with an optimization-guided falsification approach from the literature is provided.
Furthermore, a cost function that guides the test generation toward almost-avoidable collisions or near-misses is proposed.
\end{abstract}

\input{rrt_approach}
\input{rrt_approach_conclusions}


\bibliographystyle{eptcs}
\bibliography{rrtbased_bib.bib}

\end{document}

%% file: rrt_approach.tex
\section{Introduction} \label{sec:rrt_approach}
Autonomous vehicles are safety-critical systems, and they should be tested thoroughly before they are deployed on public roads.
Although the testing in real traffic environments is not fully replaceable, simulation-based testing provides many advantages such as fully controllable environments, ground-truth information, ability to try a massive number of scenarios, and creating risky scenarios without risking human life or the vehicle under test.

Optimization-guided falsification techniques utilize optimization engines to generate challenging scenarios for an Autonomous Vehicle (AV) under test with the ultimate goal of finding a scenario in which the Vehicle Under Test~(VUT), also called as \textit{Ego vehicle}, fails to satisfy its requirements \cite{FainekosSUY12acc,tuncali2016itsc,tuncali2018iv,DreossiEtAl2017rmlw}.
In our previous work \cite{tuncali2016itsc}, we have used an optimization-guided falsification tool, \staliro \cite{FainekosSUY12acc}, to search for maneuvers of agent actors (other road occupants) with the aim of identifying the boundary between the safe and unsafe operation of VUT.

\begin{figure}[htp]
	\begin{centering}
		\includegraphics[trim={0in 8.5in 1.5in 0in}, clip,width=\columnwidth]{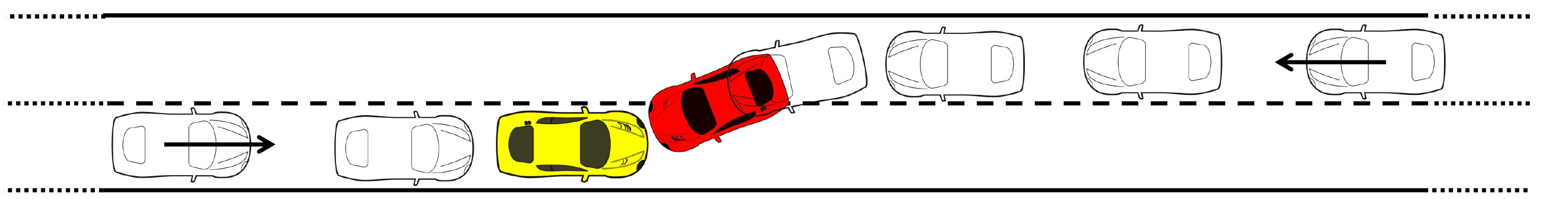}
		\caption[An unavoidable collision example.]{An unavoidable collision example. Red car suddenly moves into the Ego vehicle's lane at a very short distance.}
		\label{fig:unavoidable_collision}
	\end{centering}
\end{figure}

In this paper, we propose an alternative approach to the falsification-based approach that is described in \cite{tuncali2016itsc}.
We consider the test generation problem as a path planning problem for agent actors with the aim of creating interesting collisions with vehicles under test by utilizing Rapidly-exploring Random Trees~(RRTs).
Because of the nature of the driving environment, an automated vehicle cannot avoid all collisions.
For instance, a collision with a vehicle that loses control and drives into the automated vehicle may not be avoidable.
\figg{\ref{fig:unavoidable_collision}} illustrates an unavoidable collision example.
However, identifying the boundaries between the collisions that are barely avoided or the collisions that could have been avoided with minor changes in the control or environment would be valuable for the engineering teams to improve the safety-related capabilities of the system.
We focus on finding test cases that lead to behaviors at the proximity of the boundary between collisions and near-collisions.
For this purpose, we propose a cost function that can guide the test generation toward that boundary.

RRTs have been first developed for robot path/motion planning problems \cite{lavalle1998rapidly,jaillet2008transition,karaman2011sampling}.
However, thanks to their ability to efficiently search over high-dimensional spaces, RRT-based approaches also deliver promising results in the test generation domain \cite{esposito2004adaptive,kim2005rrt,branicky2006sampling,dang2008sensitive,plaku2009hybrid,DreossiDDKJD15nfm}.

Since their first introduction, many variants of RRTs have been proposed.
In \cite{jaillet2008transition}, a method called Transition-based RRT~(T-RRT) was introduced.
Transition-based RRT method extends the classical RRT by incorporating additional cost criteria to the explored paths rather than only aiming to reach a target configuration.
T-RRT borrows the notion of transitions tests from stochastic optimization approaches.
Hence, it can be considered as a method that is merging RRTs with stochastic optimization.
Furthermore, T-RRT controls exploration versus refinement using a method called \textit{minimal expansion control} which helps to promote the expansion of a tree to the unexplored areas of the search space.
Efficiency of T-RRT on continuous cost spaces is studied in \cite{jaillet2008transition}.
An optimal RRT approach, RRT*, was proposed, and the optimality was analyzed by Karaman and Frazzoli in \cite{karaman2011sampling}.

\section{Overview of the Approach}
In our approach, we utilize notions from RRT* \cite{karaman2011sampling} and T-RRT \cite{jaillet2008transition} with a custom cost function that we propose to find boundary-case collisions.
We implement our version of \textit{minimal expansion control} using the notion of \textit{sparseness} from evolutionary algorithms that perform novelty search \cite{stanley2002evolving,lehman2008exploiting}.

With this approach, which we will refer to as ``RRT-based approach'', we address the following limitations of our falsification-based approach \cite{tuncali2016itsc}:
\begin{itemize}
	\item [1.] In \cite{tuncali2016itsc}, we use a limited number of control points over the longitudinal position axis as the specific points where the lateral axis of the vehicle trajectories are sampled.
	As the number of control points decreases, possible variations in shapes of the trajectories are limited.
	On the other hand, increasing the number of control points also increases the dimension of the search space, which makes the problem more challenging.
	\item [2.] In \cite{tuncali2016itsc}, the duration of the simulations is fixed.
	With the RRT-based approach, although there is an inherent limit on the maximum simulation duration that is dictated by the maximum number of RRT nodes, there is flexibility in the duration of the simulations.
	So, non-promising simulations are stopped earlier while more promising ones can be executed for longer times.
	\item[3.] With the RRT-based approach, we can minimize the need for hand-designing a test scenario in detail and allow more freedom in the exploration of the space compared to the falsification-based approach.
	\item[4.] RRT-based approach is promising to avoid local minima that can be challenging to the falsification-based approach proposed in \cite{tuncali2016itsc}.
\end{itemize}

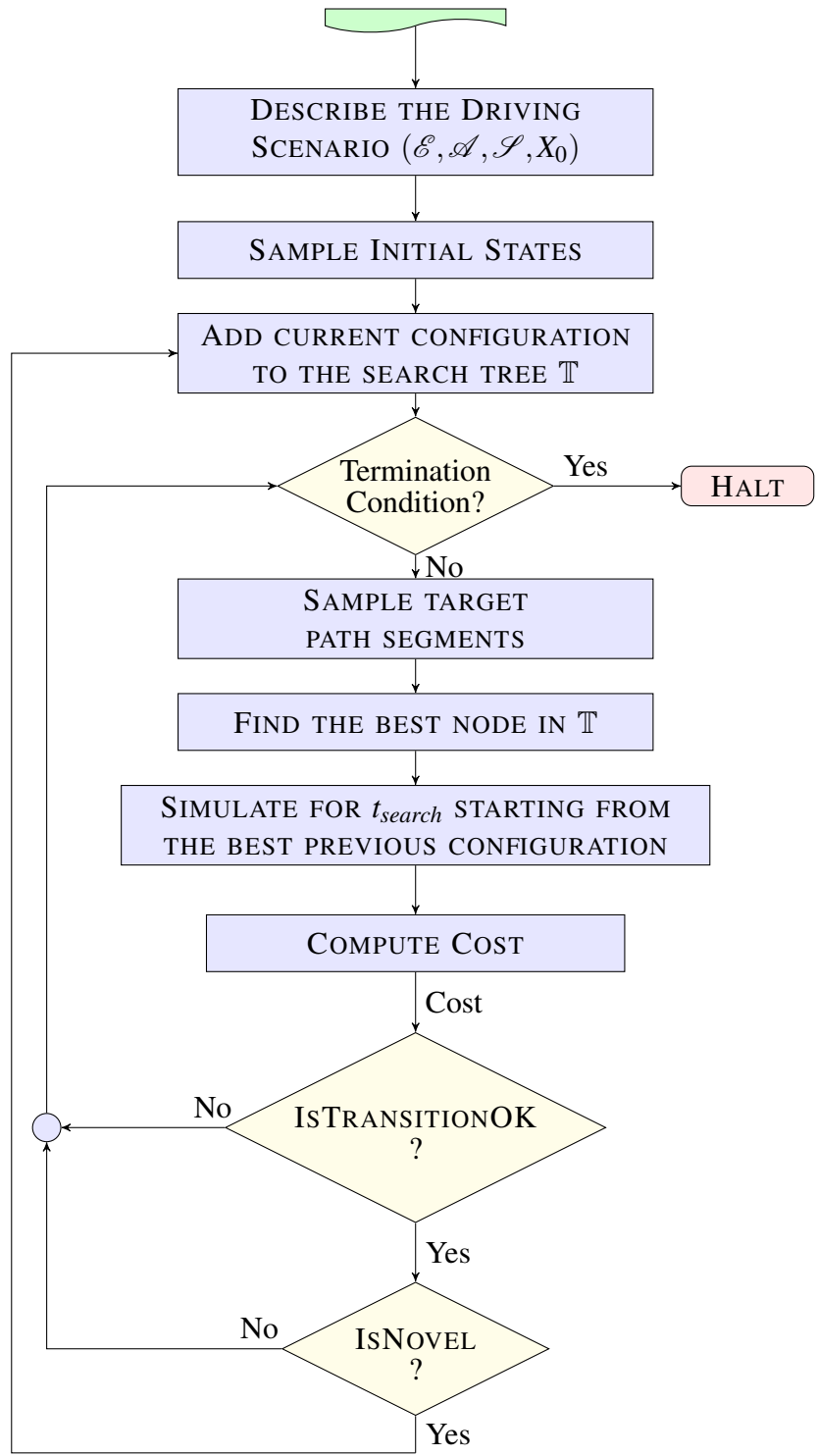
\begin{figure}[htp]
	\begin{centering}
		\input{RRT_approach_flowchart}
		\caption{Flowchart illustrating the RRT-based approach.}
		\label{fig:rrt_approach_flowchart}
	\end{centering}
\end{figure}

The flowchart of our RRT-based approach is shown in \fig{\ref{fig:rrt_approach_flowchart}}.
In the rest of this section, we will describe the key components of our approach.

\subsection{Initializing the Search}
A scenario, $\drivingScenario$, is described by the sets of Ego vehicles $\setEgo$, agent actors $\setAgent$ and environment $\setSurroundings$ as $\drivingScenario = \setEgo \cup \setAgent \cup \setSurroundings$.
The model of the driving scenario is given as $\Model_\drivingScenario = (\stSet_\drivingScenario, \inpSet_\drivingScenario,\prSet_\drivingScenario,\sim_\drivingScenario)$ where $\stSet_\drivingScenario$ is the set of system states, $\inpSet_\drivingScenario$ is the set of system inputs, $\prSet_\drivingScenario$ is the set of system parameters, and $\sim_\drivingScenario$ is the system simulation function, respectively, for the overall scenario.
Given $\stVec \in \stSpace$, $\stVecNext=\sim(\stVec,\inpVec,\prVec,t)$ is the state reached starting from the system state $\stVec$ after time $t\in \timeSet$ under the input vector~$\inpVec \in \inpSpace$ and the parameter vector $\prVec \in \prSpace$.
Hence, the system simulation function is a mapping $\sim :\stSpace \times \inpSpace \times \prSpace \times \timeSet \rightarrow \stSpace$ where $\timeSet$ is a discrete set of time samples $t_0, t_1, \ldots, t_N$, with $t_i<t_{i+1}$. 

In this approach, the set of inputs, $\inpSet_\drivingScenario$, can simply be a set of target paths for the agent actors, as well as inputs to the other entities in the simulation environment such as models of road and weather conditions.
After general outlines of the driving scenario are described, the sampling space for the initial states, \ie $\stSpace_{0,\drivingScenario}$, is used to sample initial configurations of the simulation entities, including Ego vehicles and agent actors.

\subsection{Information Stored on RRT Tree Nodes}
A tree grows while seeking to discover interesting behaviors.
While growing the tree, instead of executing simulation traces starting from the initial configuration, only a partial simulation is executed starting from an existing node in the tree.
For that purpose, the state of the system, the state of the controllers, and the simulation time are stored on the tree nodes.
\tbl{\ref{table:rrt_nodes_data}} provides more details on the data stored on the tree nodes after each partial simulation.
The history-related fields will be empty for the root nodes of trees.

\begin{table*}[htp] 
	\centering
	\begin{tabularx}{\textwidth}{rX}
		\hline
		\textbf{Data} & \textbf{Description}\\ \hline
		State & The state of the system at the end of the corresponding part of the simulation is stored. It serves as the initial state for a new partial simulation that is starting from the configuration represented in the current node.\\
		State History & (optional) The state history over the corresponding part of the simulation is used to (i) reproduce agent behaviors after the search is over (ii) to simulate any sensor delays for the next simulation step.\\
		Input History & (optional) The history of inputs to the system is used as the past input data for the next simulation step, which may be useful for the systems that need to remember past inputs. This data is also valuable for analysis purposes when the search is over.\\
		Controller State & The final state of controllers, if available, are stored so that the next step of the simulation can be started from the current node without needing to run the whole simulation from the starting node to the current node.\\
		Simulation Time & The time at the end of the corresponding part of the simulation is stored.\\\hline
	\end{tabularx}
	\caption{Data stored on the RRT nodes.}
	\label{table:rrt_nodes_data}
\end{table*}

\subsection{Sampling a Target Path Segment}
\begin{figure}[htp]
	\begin{centering}
		\includegraphics[trim={0 6.7in 7.25in 0.5in}, clip,width=0.75\columnwidth]{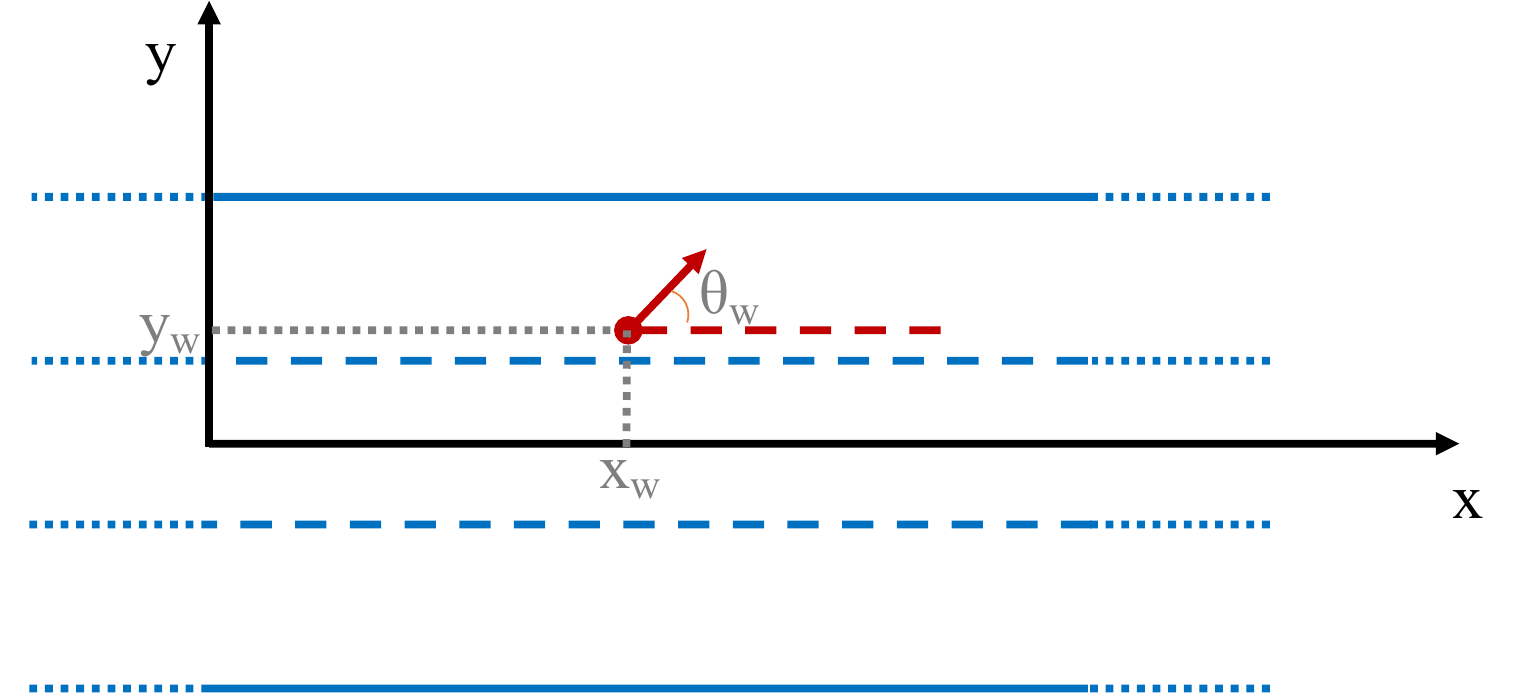}
		\caption{Sampling a waypoint.}
		\label{fig:rrt_wpt_sampling}
	\end{centering}
\end{figure}

A sample target path segment is simply a set of waypoints which is used as an immediate target for the corresponding vehicle.
A waypoint is denoted as $\waypoint = (x_\waypoint, y_\waypoint, \waypointangle_\waypoint, v_\waypoint) \in \prSpace_{\waypoint}$ where $x_\waypoint, y_\waypoint, \waypointangle_\waypoint, v_\waypoint$ are the $x$-coordinate, $y$-coordinate, target driving direction and the target speed at the waypoint. 
The sampling space for the waypoints is defined by a corresponding parameter space $\prSpace_{\waypoint}= \prSpace_{\waypoint,x} \times \prSpace_{\waypoint,y} \times \prSpace_{\waypoint,\waypointangle} \times \prSpace_{\waypoint,v}$ that describe the limits on the $x-y$ coordinates, driving direction, and target speed where $\prSpace_{\waypoint} \subseteq \prSpace_{\drivingScenario}$.
An example waypoint sampled on a straight road is shown in \fig{\ref{fig:rrt_wpt_sampling}}.
A coordinate transformation can be applied for sampling from curved roads.
Although the example waypoint in \fig{\ref{fig:rrt_wpt_sampling}} is sampled from a road, the sample space of the waypoint doesn't have to be the same as the area of a road in the simulation.
It may be defined to go beyond the road limits, it may be limited to only a part of a road, or it may be completely irrelevant to a road in the simulation environment.

\begin{figure}[htp]
	\begin{centering}
		\includegraphics[trim={0 7.2in 7.3in 0}, clip,width=0.75\columnwidth]{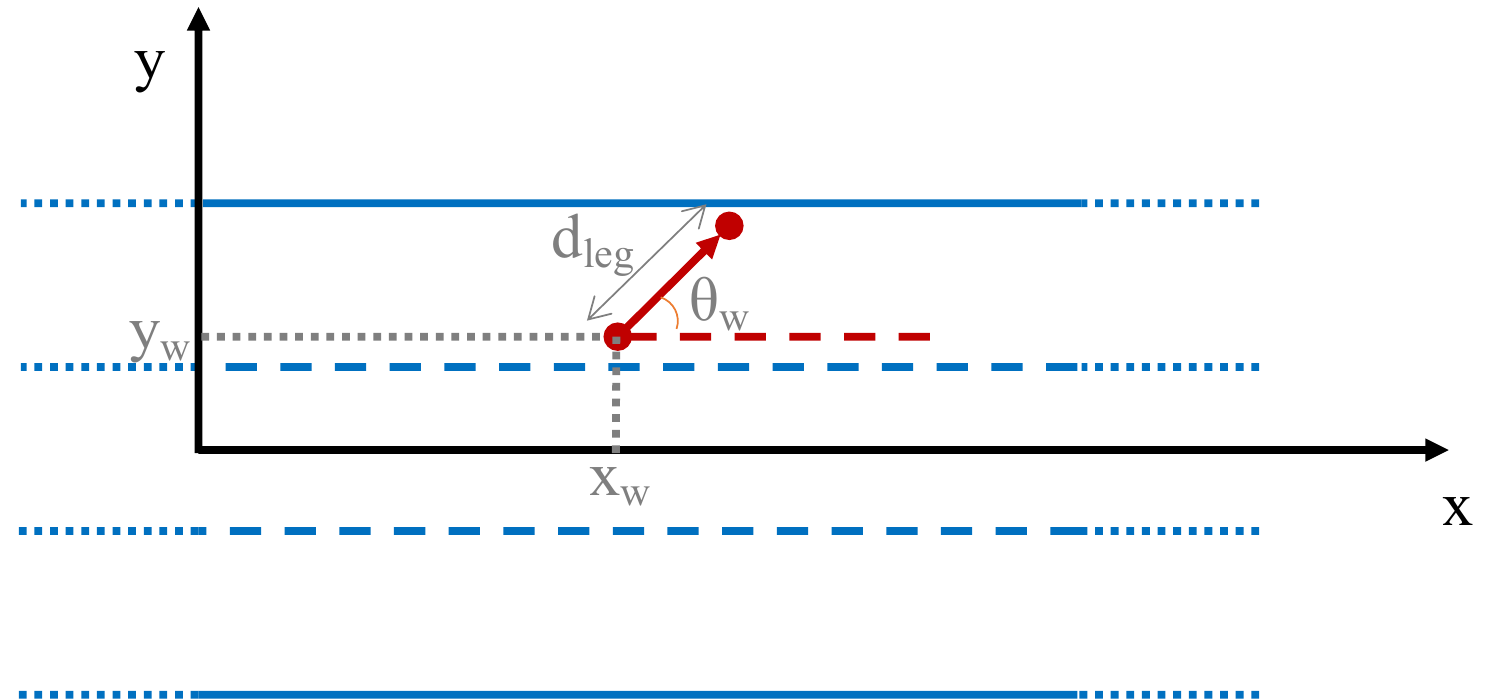}
		\caption{Sampling a target path segment.}
		\label{fig:rrt_wpt_endpoint}
	\end{centering}
\end{figure}

Once a waypoint is sampled, the next step in sampling a target path segment is to add an endpoint at a predefined distance $d_{leg}$ from the waypoint, along the direction of the waypoint.
\figg{\ref{fig:rrt_wpt_endpoint}} shows a target path segment formed using this approach.
The sampled target path segment for this example can be denoted by $\targetPath = ((x_\waypoint, y_\waypoint, \waypointangle_\waypoint,v_\waypoint),(x_{\waypoint2}, y_{\waypoint2}, \waypointangle_\waypoint, v_\waypoint))$.

\begin{figure}[htp]
	\begin{centering}
		\includegraphics[trim={0 7.2in 7.3in 0}, clip,width=0.75\columnwidth]{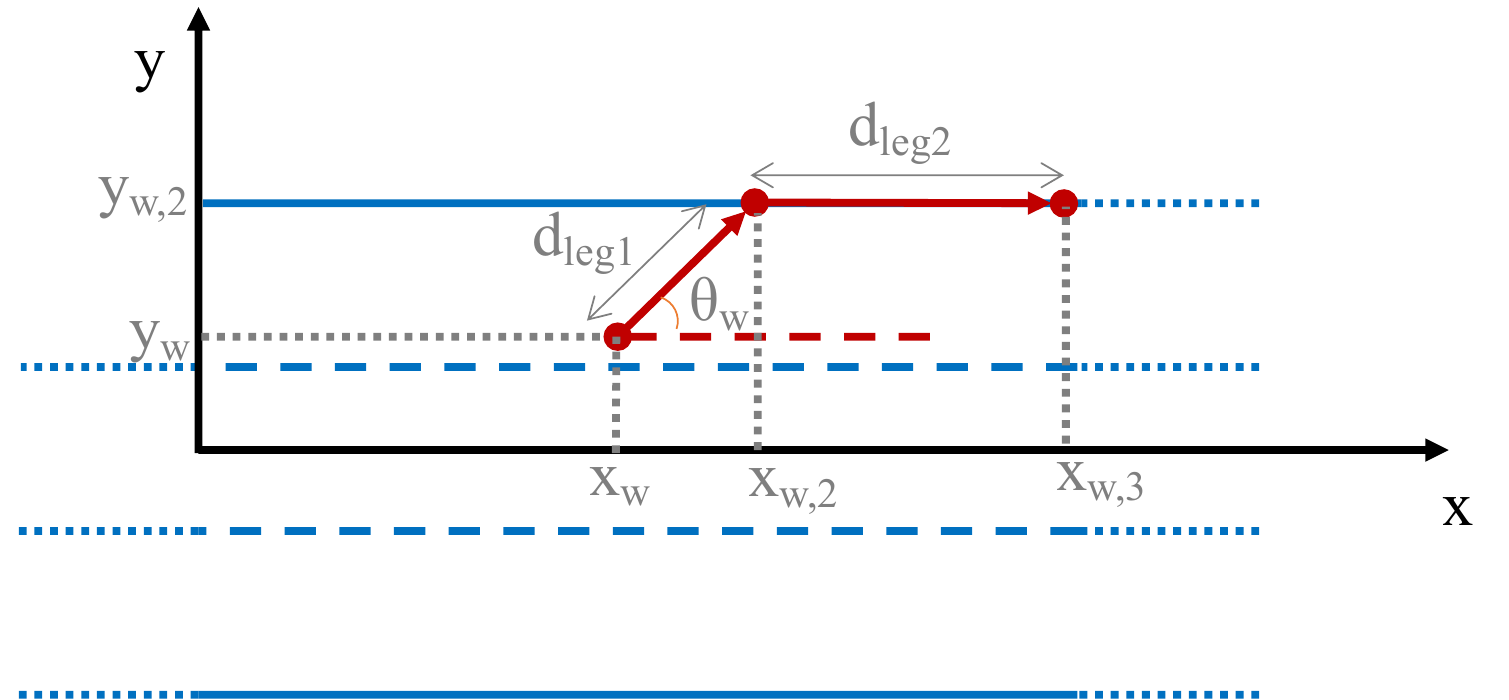}
		\caption{Sampling a target path segment with space constraints.}
		\label{fig:rrt_two_segment_path}
	\end{centering}
\end{figure}

If the endpoint of a target path segment is outside the sampling space of the waypoint, we simply break the segment at the boundary of the sampling space and add a second leg along the boundary in the direction closest to the waypoint direction.
\figg{\ref{fig:rrt_two_segment_path}} shows an example target path section for a longer $d_{leg}=d_{leg1} + d_{leg2}$ compared to the one in \fig{\ref{fig:rrt_wpt_endpoint}}.
The sampled target path segment for this example can be denoted by $\targetPath = ((x_\waypoint, y_\waypoint, \waypointangle_\waypoint, v_\waypoint),(x_{\waypoint2}, y_{\waypoint2}, \waypointangle_{\waypoint2}, v_\waypoint), (x_{\waypoint3}, y_{\waypoint3}, \waypointangle_{\waypoint2}, v_\waypoint))$.

After the shape of the target segment is decided, we also sample a target speed, $v_\waypoint$, for the target path segment.

This approach is applied to each agent vehicle for sampling their target path segments.
Note that the framework we propose allows using a different algorithm for selecting a target path and/or any other type of input to the agent vehicles.

\subsection{Selecting the Best Node from the Search Tree}
Once target path segments for the agent vehicles are sampled, we pick one node from the existing tree, as the initial configuration for the simulation that will be executed with the sampled target path segments.
There is no single correct approach to decide which node of the tree would be the best choice.

In our study, the notion of selecting the optimal node is adopted from the RRT* method \cite{karaman2011sampling}.
In \cite{karaman2011sampling}, when adding a new node to the tree, existing nodes in a neighborhood of the new node are all checked and the one which minimizes the cost is selected as the previous node.
Our approach has similarities to the RRT* one.
We compute the sum of distances from each vehicle to the starting point of their target path segment with the constraint that the configurations of all vehicles are behind the start position of the initial target waypoint with respect to the driving direction of that waypoint.
Then, we execute partial simulations from the best $n$ candidate previous nodes ($n=5$ for our case studies) and pick the one which gives the minimum cost.
We believe that this approach is promising to create relatively natural-looking vehicle trajectories while still allowing enough randomness in the maneuvers.

In \cite{karaman2011sampling}, after adding the new node to the tree, there is a rewiring step which modifies the connections to the other existing nodes in the neighborhood of the new node.
The rewiring step checks whether reaching to an existing node in that neighborhood from the newly added node would result in a reduction in the cost without violating any constraints such as obstacles.
If so, it replaces the existing edge incoming to that node with an edge originating from the newly added node.
Application of the rewiring step is straightforward for path planning problems on Euclidean spaces where the tree nodes represent planned waypoints on a path instead of the simulated vehicle configurations.
In our approach, the tree nodes are the resulting configurations reached by the simulated vehicles with a controller and an input target path.
Hence, the rewiring step requires the execution of partial simulations starting from the newly added node to the other nodes in the neighborhood of the newly added node.
Since the resulting configuration will most likely be different at the end of such a partial simulation, the configurations on the target nodes will change.
This creates the necessity to execute simulations from the updated nodes to all of the remaining tree nodes that can be reached from the modified node.
Hence, the rewiring step can be computationally costly in our approach, and so, we do not apply the rewiring step and leave it as a future work for which the applicability should be analyzed.

We would like to emphasize that the function used for selecting the best previous node is user-configurable in our framework and depending on how much randomness is plausible in the generated driving paths, a different algorithm, \eg simply selecting the closest node, can be utilized.

\subsection{Simulating the System}
After obtaining a set of target path segments for agent vehicles and deciding the initial configuration for the simulation, we create the simulation scene in the simulation environment using the data stored in the selected node of the search tree.
That is, we set the initial states of the simulation entities and initialize the Ego vehicle controllers with the previous inputs and the saved controller states.
We also pass the sampled target path segments to the agent vehicles as inputs.
Finally, we simulate the system for $\searchdt$ time and collect state and input histories at each time step of the simulation.
For our setup, we use \matlab simulations, however, this is not mandatory and another simulator can be used.
Note that if the simulator and the Ego vehicle controllers allow saving the state and continuing simulation from a saved state, \textit{which is the case in our setup}, the time spent in the simulations can be radically reduced because it would be enough to simulate only the new part of the simulation.
Otherwise, the simulation should always start from the root node of the tree and run until the current target time.

\subsection{Cost Function}
After a simulation is executed, a cost function is used to compute how close the simulation trace is to an interesting behavior.
The approach we describe here can be utilized to discover other types of interesting/failing behavior; however, our target in this work is to explore the behaviors that are on the boundary between safe and unsafe operation.
Hence, an interesting behavior for our purposes would be (i) a collision between an Ego vehicle and an agent that could have been avoided with a minor change in the control applied or agent trajectories, (ii) an almost-collision (near-collision) which could have ended with a collision with a minor change in the control applied or agent trajectories.

The properties of a good cost function that would guide the search toward an interesting behavior for our purposes can be listed as follows:
\begin{itemize}
	\item Among two similar collisions between an Ego vehicle and an agent vehicle, the one which has the smaller magnitude in the relative speed between the vehicles should have a smaller cost, as a smaller change in the speed of the Ego vehicle would be enough to avoid the collision.
	\item Among two similar collisions, the one which has the smaller impact area, \ie the area of the collision surface, should have a smaller cost, as a smaller change in the steering maneuver of the Ego vehicle would be enough to avoid the collision.
	\item For vehicle paths without a collision, a smaller time-to-collision at any point of the path, and a smaller corresponding collision speed and a smaller area for that collision-to-be should lead to a smaller cost.
\end{itemize}

\begin{figure}[htp]
	\begin{centering}
		\includegraphics[trim={0.7in 2.75in 0.75in 3in}, clip,width=0.8\columnwidth]{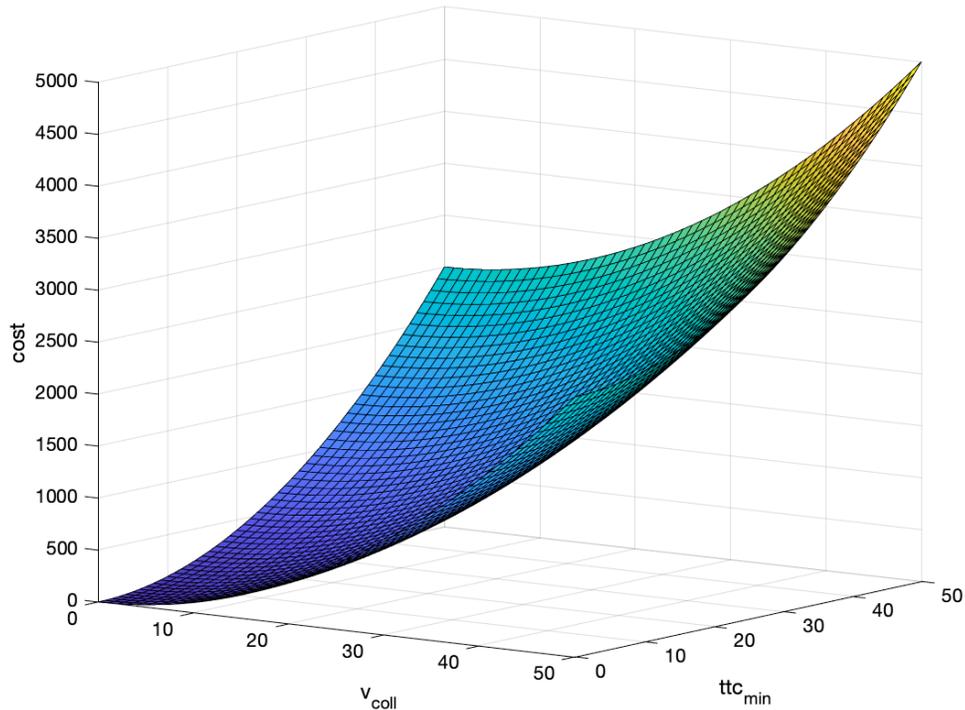}
		\caption{Cost function to guide the search toward the boundary between collisions and near collisions.}
		\label{fig:newcostfunc}
	\end{centering}
\end{figure}

We propose the following cost function:
\begin{equation}
\costFun(\outTraj)= (1+\collisionSurface)(v_{coll,\outTraj}^2 + ttc_{min,\outTraj}^2)
\label{eqn:collisioncost_new}
\end{equation}
where $\collisionSurface \in [0,1]$ is the ratio of the collision surface to the overall surface on the collision side of the vehicle, $v_{coll,\outTraj}$ is the relative speed of the vehicles at the moment of collision, and $ttc_{min,\outTraj}$ is the mimimum time-to-collision encountered during the simulation output trace $\outTraj$.
For the simulations with a collision, $ttc_{min,\outTraj}$ is $0$.
For the simulations without a collision, $\collisionSurface$ and $v_{coll,\outTraj}$ are computed at the instance of smallest time-to-collision with the assumption that the vehicles continue their motion without changing their speeds and orientations.
When the simulation output trace $\outTraj$ contains collision(s) with Ego vehicle, we only consider the first collision of an Ego vehicle with any object for computing \eqn{\ref{eqn:collisioncost_new}}.
\figg{\ref{fig:newcostfunc}} shows the function with respect to the minimum time-to-collision and collision speed variables for a fixed collision surface.
The effect of the collision surface to the cost is linear.

\subsection{Transition Check Function}
The function we use for accepting a new configuration based on the cost is similar to the one proposed in \cite{jaillet2008transition}.
\algo{\ref{alg:transitionTest}} repeats it for convenience.
We denote the newly simulated configuration as the \textit{candidate} and the initial configuration selected from the search tree as \textit{previous} configuration.
In the algorithm, $K$ is a constant parameter normalizing the change in the cost, $T$ is the temperature parameter that is governing the likelihood of acceptance. 
The temperature parameter $T$ is adaptively tuned by multiplying with or dividing by $\alpha$ with respect to the ratio of rejections to acceptances.

\begin{algorithm}[htp]
	\caption{Algorithm used to check acceptance of a new configuration based on the change in the cost.
		\label{alg:transitionTest}}
	\begin{algorithmic}[1]
		\Function{IsTransitionOK}{$c_{prev}, c_{cand}$} \Comment $c_{prev}$ and $c_{cand}$ are the costs associated with the previous and candidate configurations, respectively. $\alpha$ and $K$ are constant parameters, and $T$ is a persistent parameter.
		\If {$c_{cand} < c_{prev}$}
		\State \Return{True}
		\EndIf
		\If {$rand(0,1) < e^{(c_{prev}-c_{cand})/(K*T)}$}
		\State $T=T/\alpha$
		\State \textit{numberOfFails} = 0
		\State \Return {True}
		\Else
		\If {\textit{numberOfFails} > \textit{maxNumberOfFails}}
		\State $T=T*\alpha$
		\State \textit{numberOfFails} = 0
		\Else
		\State \textit{numberOfFails} = \textit{numberOfFails} + 1
		\EndIf
		\State \Return{False}
		\EndIf
		\EndFunction
	\end{algorithmic}
\end{algorithm}

\subsection{Novelty Function}
To get better coverage of the state space and to avoid local minima, we reward novelty in our search.
For an Ego vehicle $\egoVhc_i\in\setEgo$ and an agent $\agent_j\in\setAgent$, we define $\stVec_{\egoVhc_i, \agent_j}[k]$ as the vector of relative states and change in relative states at discrete simulation time $k\in[0,n]$, \ie $\stVec_{\egoVhc_i, \agent_j}[k] = (\stVec_{\egoVhc_i}[k] - \stVec_{\agent_j}[k], (\stVec_{\egoVhc_i}[k] - \stVec_{\agent_j}[k]) - (\stVec_{\egoVhc_i}[k-1] - \stVec_{\agent_j}[k-1]))$.

We compute the novelty of $\stVec_{\egoVhc_i, \agent_j}[k]$ as follows:
\begin{equation}
\novelty = \sum_{l=0}^{m}dist(\stVec_{\egoVhc_i, \agent_j}[k], \mu_l)
\label{eqn:rrt_novelty}
\end{equation}
where $\mu_l \in \stSet_{rel,k-1}$ is the $l^{th}$ nearest neighbor of $\stVec_{\egoVhc_i, \agent_j}[k]$ in the set $\stSet_{rel,k-1}$ which contains $\stVec_{\egoVhc_i, \agent_j}$ vectors for all ego-agent pairs for all times before $k$.
The function $dist$ computes the Mahalanobis distance between $\stVec_{\egoVhc_i, \agent_j}[k]$ and the elements of its $m$-nearest neighbors set.
We choose to use the Mahalanobis distance because of its ability to provide a dissimilarity measure between two observations by utilizing the sample covariance matrix \cite{de2000mahalanobis}.

As each new configuration has a corresponding partial simulation of length $\searchdt$ starting from a previous configuration, we compute the novelty for the trace of that partial simulation using \algo{\ref{alg:noveltyCheck}}.

\begin{algorithm}[htp]
	\caption{Algorithm used to check the novelty of a new configuration.
		\label{alg:noveltyCheck}}
	\begin{algorithmic}[1]
		\Function{IsNovel}{$\stTraj,\setEgo,\setAgent,k_{start},k_{end},c_{prev}, c_{cand}$}
		\State $\mathbf{N_{last}}$ is persistent and keeps the last computed $10$ novelty values. \textbf{numR} is persistent and keeps the number of rejections. \textbf{maxReject} is the maximum number of consecutive rejections. $\stSet_{rel,k}$ is persistent and keeps the set of all past relative state computations. $c_{prev}$ and $c_{cand}$ are the costs associated with the previous and candidate configurations, respectively.
		\State \ 
		\State Initialize $\mathbf{N}$ as an empty set
		\For {each $\egoVhc_i \in \setEgo$}
		\For {each $\agent_j \in \setAgent$}
		\For {$k = k_{start}$ to $k_{end}$}
		\State Compute $\stVec_{\egoVhc_i, \agent_j}[k]$ from $\stTraj$
		\State Compute $m$-nearest neighbors of $\stVec_{\egoVhc_i, \agent_j}[k]$ in $\stSet_{rel,k-1}$
		\State Compute $\novelty$ (novelty) with \eqn{\ref{eqn:rrt_novelty}}
		\State Add $\novelty$ to $\mathbf{N}$
		\State Add $\stVec_{\egoVhc_i, \agent_j}[k]$ to $\stSet_{rel,k}$
		\EndFor
		\EndFor
		\EndFor
		\State $\novelty = \max(\mathbf{N})$
		\State Update $\mathbf{N_{last}}$ to keep the last computed $10$ novelty values
		\If {$c_{cand} < 0.9c_{prev}$ \textbf{or} numR > maxReject \textbf{or} $|\mathbf{N_{last}}|<10$ \textbf{ or} $\novelty>mean(\mathbf{N_{last}})$}
		\State numR = 0
		\State \Return {True}
		\Else
		\State numR = numR + 1
		\State \Return {False}
		\EndIf
		\EndFunction
	\end{algorithmic}
\end{algorithm}

\subsection{Termination Condition}
Our algorithm checks a set of termination conditions to stop the search and returns the configuration which has the minimum cost associated with it.
One of the termination conditions we use is a threshold for the minimum interesting cost.
Another termination condition is a preset maximum overall time spent.
Alternative termination conditions can be used, \eg a maximum number of nodes in the search tree.

\section{Case Studies}
Here, we present 2 case studies and compare our RRT-based approach with our falsification-based approach \cite{tuncali2016itsc}.

\subsection{Case Study 1}
\subsubsection{Scenario Setup}

\begin{figure}[htp]
	\begin{centering}
		\includegraphics[trim={0in 6.75in 1.1in 0in}, clip,width=\columnwidth]{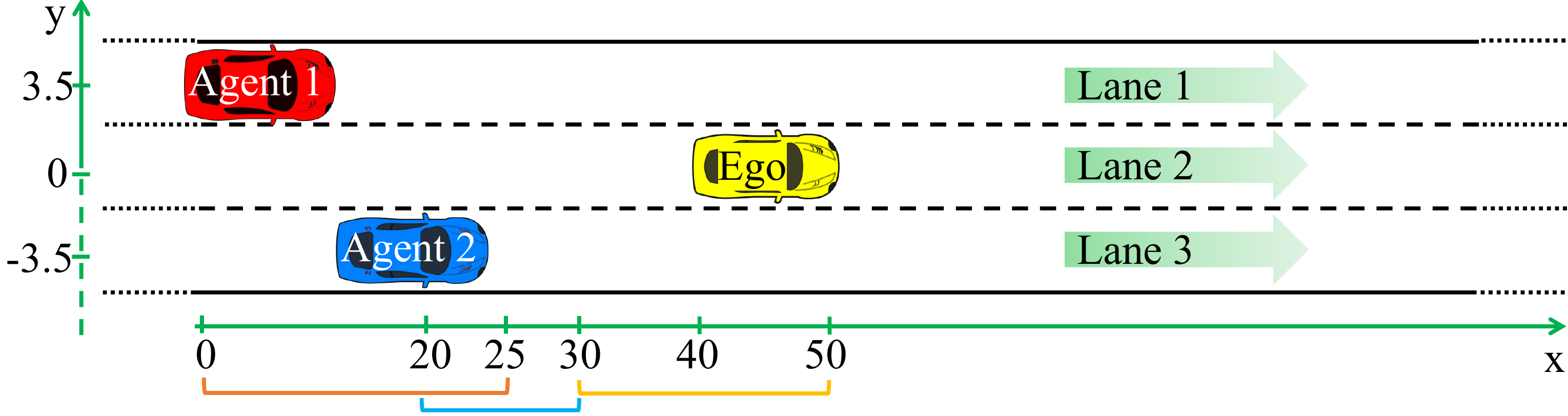}
		\caption{Vehicle initial states in the simulation setup for Case Study 1.}
		\label{fig:rrt_sample_space}
	\end{centering}
\end{figure}

In this case study, we have 2 agent vehicles and 1 Ego vehicle on a 3-lane straight road, \ie $\setAgent = \{\agent_1,\agent_2\}$ and $\setEgo = \{\egoVhc\}$.
\figg{\ref{fig:rrt_sample_space}} gives an high-level overview of our simulation setup.
The initial position of agent vehicle 1 on the $x$ axis is randomly sampled between \SI{0}{\meter} and \SI{25}{\meter}, the initial $x$ position of agent vehicle 2 is randomly sampled between \SI{10}{\meter} and \SI{20}{\meter}, and the initial $x$ position of Ego vehicle is randomly sampled between \SI{30}{\meter} and \SI{50}{\meter}.
The initial positions of agent vehicles on the $y$ axis are sampled between the centers of lane 1 and lane 3, \ie between \SI{-3.5}{\meter} and \SI{3.5}{\meter}, and the initial $y$ position of Ego vehicle is randomly sampled between \SI{-1.75}{\meter} and \SI{1.75}{\meter}, that is the lane markings separating Lane 3 and Lane 1 from Lane 2, respectively.
The initial orientation of the Ego vehicle with respect to the $x$ axis is randomly sampled between $\SI{-\pi/8}{\radian}$ and $\SI{\pi/8}{\radian}$.
The initial speed of Ego vehicle is sampled between $\SI{10}{\meter/\second}$ and $\SI{15}{\meter/\second}$ while the target speed is fixed to $\SI{15}{\meter/\second}$.
The initial speeds of the agent vehicles are sampled between $\SI{0}{\meter/\second}$ and $\SI{15}{\meter/\second}$, and their target speeds at each waypoint are sampled between $\SI{0}{\meter/\second}$ and $\SI{30}{\meter/\second}$.

\begin{figure}[htp]
	\begin{centering}
		\includegraphics[trim={1in 1.5in 0.9in 1.5in}, clip,width=0.9\columnwidth]{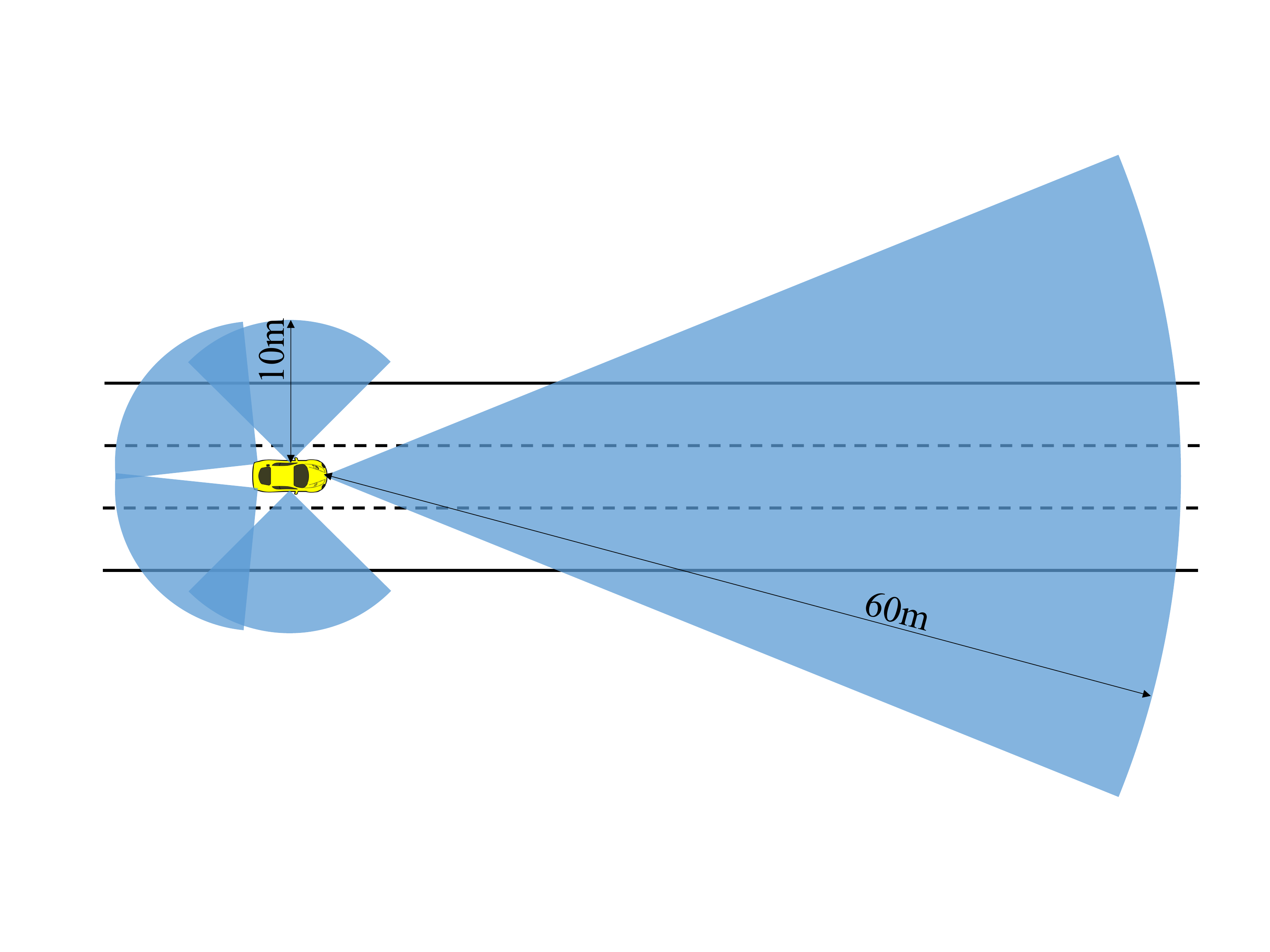}
		\caption{Ego vehicle sensor setup for Case Study 1.}
		\label{fig:rrt_sensor_setup}
	\end{centering}
\end{figure}

Ego vehicle has 5 sensors.
\figg{\ref{fig:rrt_sensor_setup}} visualizes the sensor placement and ranges of the sensors.
A long-range sensor with a \ang{45} field of view and \SI{60}{\meter} range is placed at the front of the vehicle.
Two \SI{10}{\meter}-range sensors with \ang{90} field-of-view are placed on the sides, facing left and right.
Two \SI{10}{\meter}-range sensors with \ang{90} field-of-view are placed at the rear-left and rear-right corners with an angle to scan the area behind the rear corners of the vehicle.

Agent vehicles are controlled by the \textit{move-to-pose} controller described in \cite{corke2017robotics}.
Ego vehicle controller has multiple modes based on the collision risk.
When there is no estimated collision risk, a proportional control is applied to track target driving speed.
If a collision is estimated in front, emergency brakes are applied and at the same time, depending on the occupancy of rear-left and rear-right areas, left or right steering is applied, respectively.
If a collision is estimated in front-left (right), emergency brakes are applied and if rear-right (left) area is empty, also steering is applied to the right (left).
If a collision is estimated in rear-left (right), if front and front-right (left) areas are empty, vehicle is accelerated with a right (left) steering, if only front area is empty and the vehicle on the front-right (left) is not imposing a risk, vehicle is accelerated without steering, otherwise emergency brakes are applied and right (left) steering is applied if the rear-right (left) area is empty.
If the area to which a maneuver is being done gets occupied during the maneuver, emergency brakes are applied.
Control switches back to normal mode if there is no more collision risk and a predefined time has passed since the last collision estimation.
The collision avoidance algorithm presented here is very simple and it is not comparable to a controller that could be found in a real ADS.
However, since our target in this work to study test generation approaches, rather than proposing a controller, we argue that this naive control approach is satisfactory for the purpose of this work.
For the lateral control, we use the Stanford Racing Team's controller that was used in the DARPA Grand challenge \cite{hoffmann2007autonomous}.

\subsubsection{Experiment Results}
\begin{figure}[htp]
	\begin{centering}
		\includegraphics[trim={1.2in 3.0in 1.85in 3.8in},clip,width=\columnwidth]{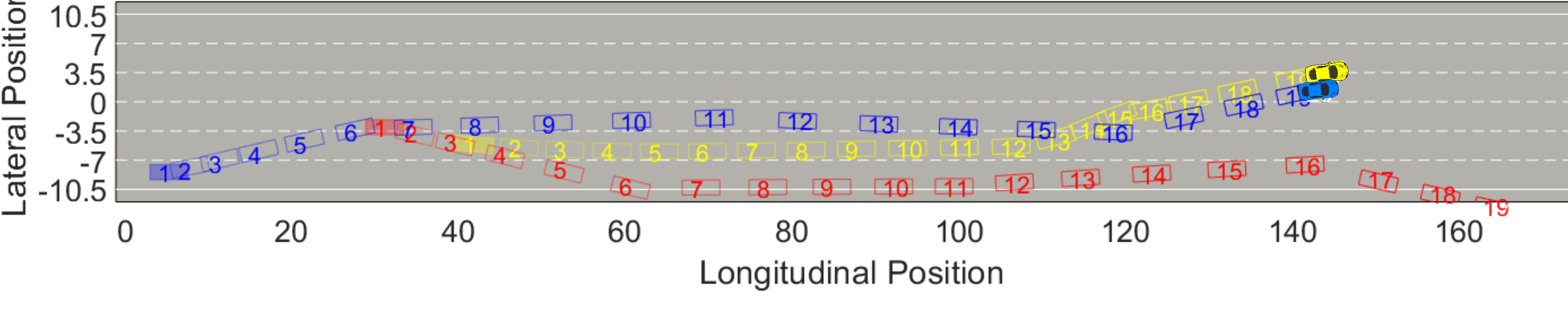}
		\caption{The minimum cost result returned by the RRT-based approach in Case Study 1.}
		\label{fig:rrt_collision_trace_case_1}
	\end{centering}
\end{figure}

\begin{figure}[htp]
	\begin{centering}
		\includegraphics[trim={0.4in 2.0in 2.4in 4.0in},clip,width=\columnwidth]{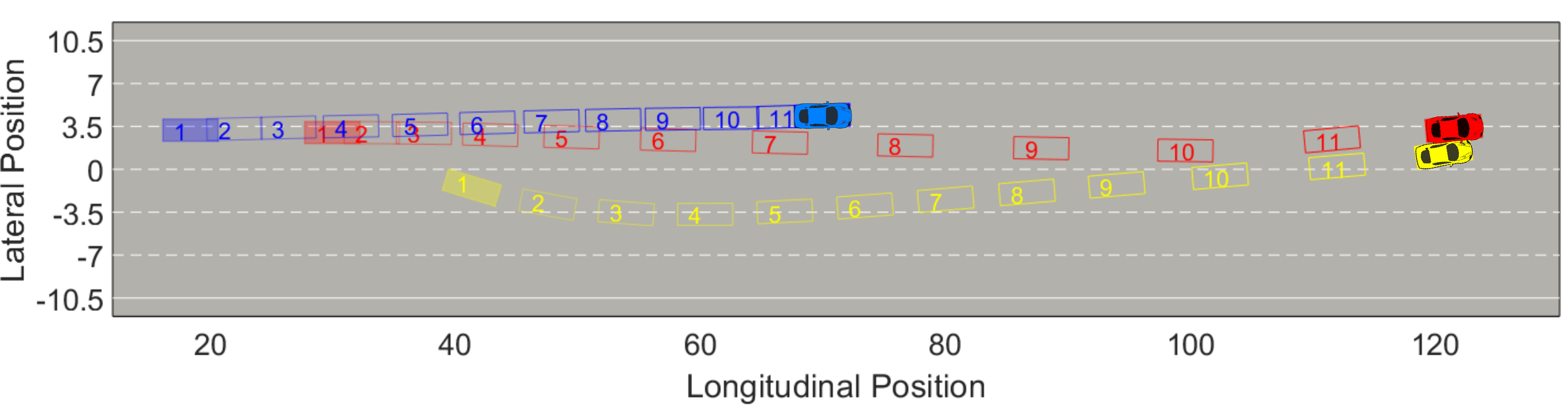}
		\caption{The minimum cost result returned by the falsification-based approach in Case Study 1.}
		\label{fig:fals_collision_trace_case_1}
	\end{centering}
\end{figure}

\begin{figure}[htp]
	\begin{centering}
		\includegraphics[trim={1.6in 3.55in 1.85in 3.6in}, clip,width=\columnwidth]{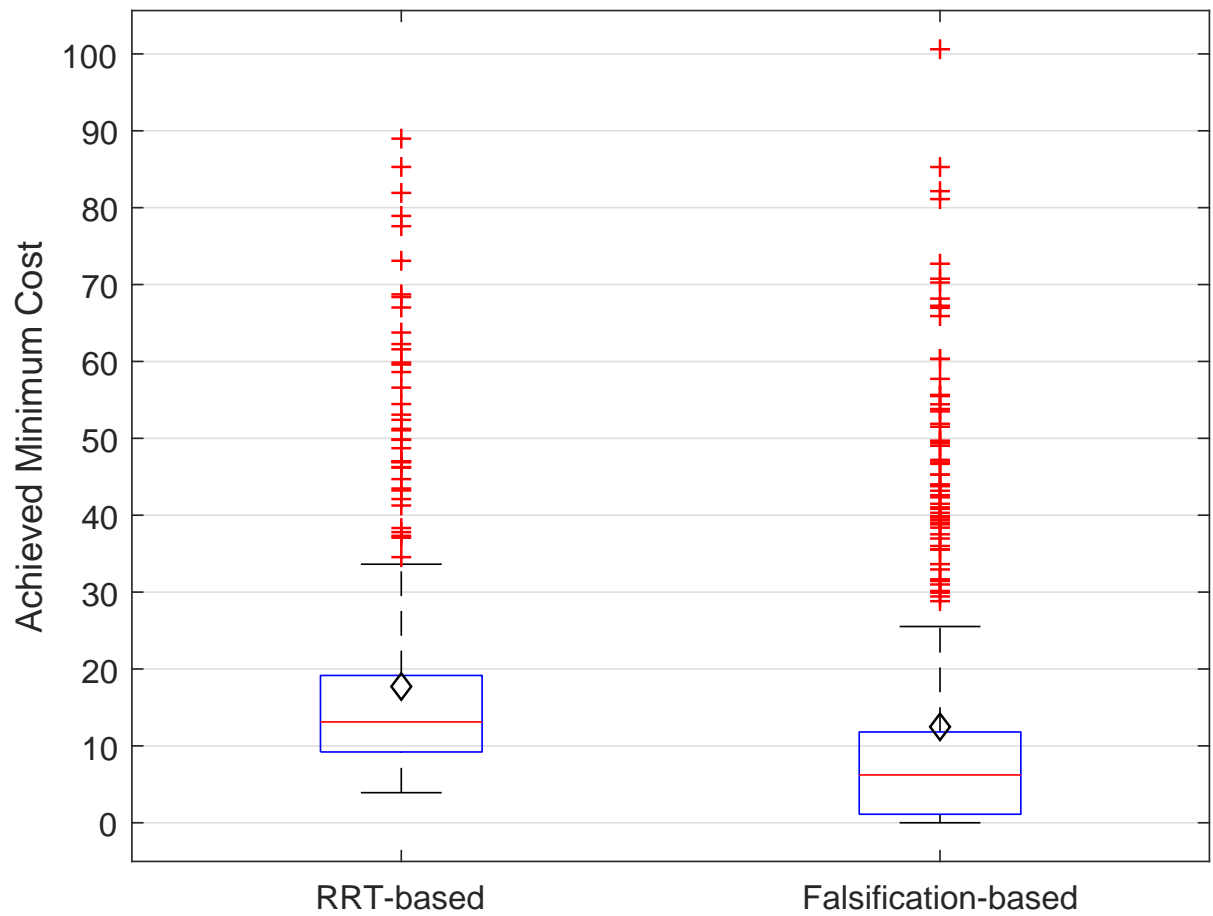}
		\caption{Comparison of the minimum cost achieved by the falsification approach and the RRT-based approach in Case Study 1.}
		\label{fig:boxplots_case_study1}
	\end{centering}
\end{figure}

We have run $200$ experiments with the falsification approach and with the RRT-based approach.
The minimum, mean and maximum costs achieved by the falsification approach were $0.0001, 12.4794$, and $100.6082$, respectively.
The minimum, mean and maximum costs achieved by the RRT-based approach were $3.9124, 17.7190$, and $88.9793$, respectively.
\figg{\ref{fig:rrt_collision_trace_case_1}} and \fig{\ref{fig:fals_collision_trace_case_1}} visualize the minimum-cost trajectories returned by the RRT-based and falsification-based approach, respectively.
Histories of the vehicles are numbered to show their evolution over time.
\figg{\ref{fig:boxplots_case_study1}} provides box and whisker diagrams of the minimum costs achieved by the two approaches among the $100$ experiments we have carried.
The black diamonds plotted on top of the box plots show the mean values for the returned minimum costs.
In this case study, the falsification-based approach achieved smaller mean cost values, as well as the smaller minimum cost compared to the RRT-based approach.
One reason for this is that, since the space between agent vehicles and Ego vehicle is open, there are not many local minimums that would make the exploration capabilities to achieve better than falsification-based approach.
As it is easy to find a trajectory that is in the neighborhood of an interesting case, falsification approach can focus on that neighborhood and minimize the cost as much as possible while RRT-based approach keeps looking for novel trajectories.

\subsection{Case Study 2}
\subsubsection{Scenario Setup}
\begin{figure}[htp]
	\begin{centering}
		\includegraphics[trim={0.2in 6in 0.2in 0.2in}, clip,width=\columnwidth]{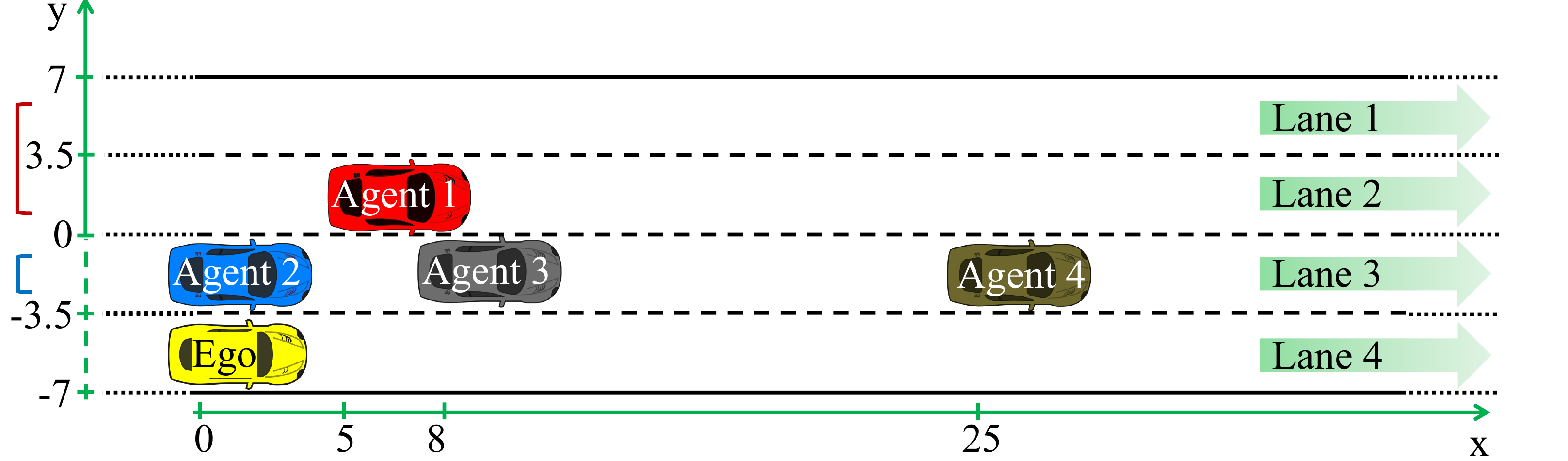}
		\caption{Initial states of the vehicles in the simulation setup for Case Study 2.}
		\label{fig:rrt_sample_space_new}
	\end{centering}
\end{figure}

In this case study, we have 4 agent vehicles and 1 Ego vehicle on a  multiple-lane straight road, \ie $\setAgent = \{\agent_1,\agent_2,\agent_3,\agent_4\}$ and $\setEgo = \{\egoVhc\}$.
\figg{\ref{fig:rrt_sample_space_new}} gives an high-level overview of our simulation setup.
The initial position of agent $\agent_1$ on the $y$ axis is randomly sampled between \SI{1.25}{\meter} and \SI{6}{\meter}, the initial $y$ positions of agent vehicles $\agent_2,\agent_3,\agent_4$ are randomly sampled between \SI{-2.25}{\meter} and \SI{-1.75}{\meter}.
The initial speed of $\agent_1$ is randomly sampled between $\SI{5}{\meter/\second}$ and $\SI{15}{\meter/\second}$, and its target speed at each waypoint is sampled between $\SI{0}{\meter/\second}$ and $\SI{30}{\meter/\second}$.
The initial and target speeds of all other vehicles are set to $\SI{15}{\meter/\second}$.
All other initial states of the vehicles are fixed.

\begin{figure}[htp]
	\begin{centering}
		\includegraphics[trim={0.2in 6.25in 1.2in 0.1in}, clip,width=\columnwidth]{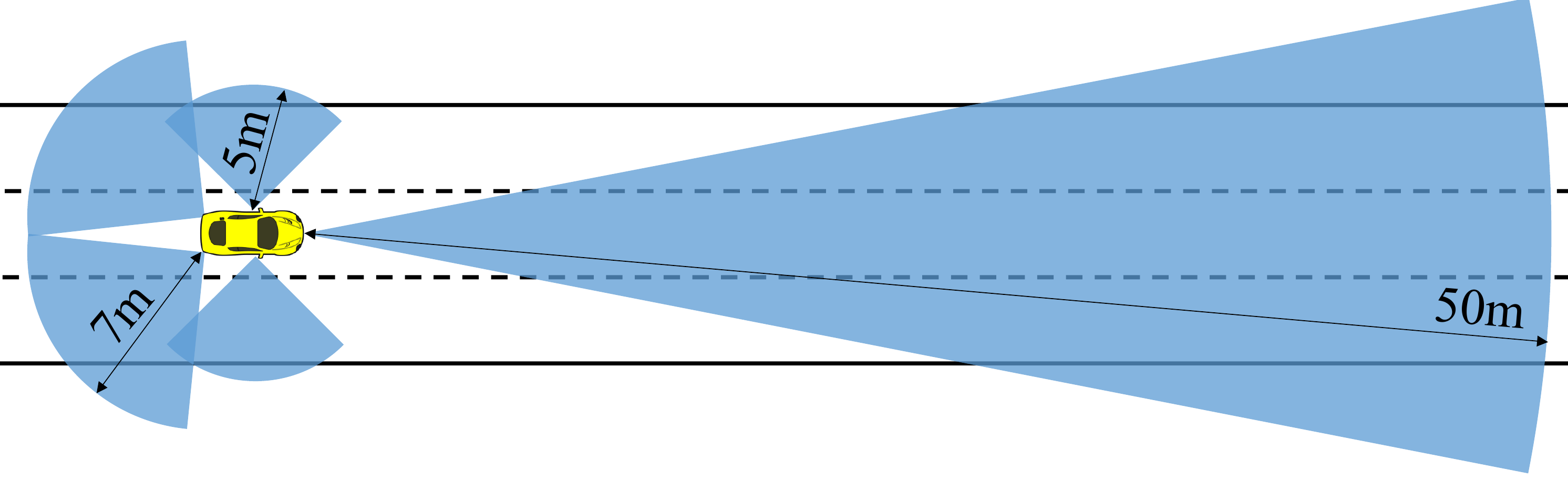}
		\caption{Ego vehicle sensor setup for Case Study 2.}
		\label{fig:rrt_sensor_setup_new}
	\end{centering}
\end{figure}

Ego vehicle has 5 sensors.
\figg{\ref{fig:rrt_sensor_setup_new}} visualizes the sensor placement and ranges of the sensors.
A long-range sensor with a \ang{22.5} field of view and \SI{50}{\meter} range is placed at the front of the vehicle.
Two \SI{5}{\meter}-range sensors with \ang{90} field-of-view are placed on the sides, facing left and right.
Two \SI{7}{\meter}-range sensors with \ang{90} field-of-view are placed at the rear-left and rear-right corners with an angle to scan the area behind the rear corners of the vehicle.

Agent vehicle $\agent_1$ is controlled by the \textit{move-to-pose} controller described in \cite{corke2017robotics}.
Agent vehicles $\agent_2,\agent_3$, and $\agent_4$ are driven with a constant speed on a straight line.
Ego vehicle controller is the same as the one described in Case Study 1.

\subsubsection{Experiment Results}
For this case study, we only search for an optimal trajectory for $\agent_1$ with the target of minimizing the cost function described in \eqn{\ref{eqn:collisioncost_new}}.
The existence of agent vehicles $\agent_2,\agent_3$, and $\agent_4$ between Ego vehicle $\egoVhc$ and agent vehicle $\agent_1$ creates many local minima for the selection of $\agent_1$ trajectories.

In this case study, we have executed $100$ experiments with both the RRT-based approach and the falsification-based approach in \matlab.
Each run of both approaches had a time-out duration of $\SI{30}{\minute}$.
Out of $100$ runs, only $5$ of the minimum-cost trajectories returned by the falsification-based approach was able to make $\agent_1$ to move into the lane of Ego vehicle, and only $2$ trajectories were able to cause a collision ($1$ high-speed collision and $1$ boundary-case collision) and, other than the collision cases,  only $1$ trajectory was able to challenge Ego vehicle by activating its collision avoidance system.
All other trajectories returned by the falsification approach were stuck in local-minima where $\agent_1$ tries to get closer to Ego vehicle and ends up colliding with one of the other agent vehicles.
On the other hand in $28$ of the minimum-cost trajectories returned by the RRT-based approach, agent $\agent_1$ was able to get into the lane of Ego vehicle and it was able to cause Ego vehicle to collide in $11$ of those cases.

\begin{figure}[htp]
	\begin{centering}
		\includegraphics[width=\columnwidth]{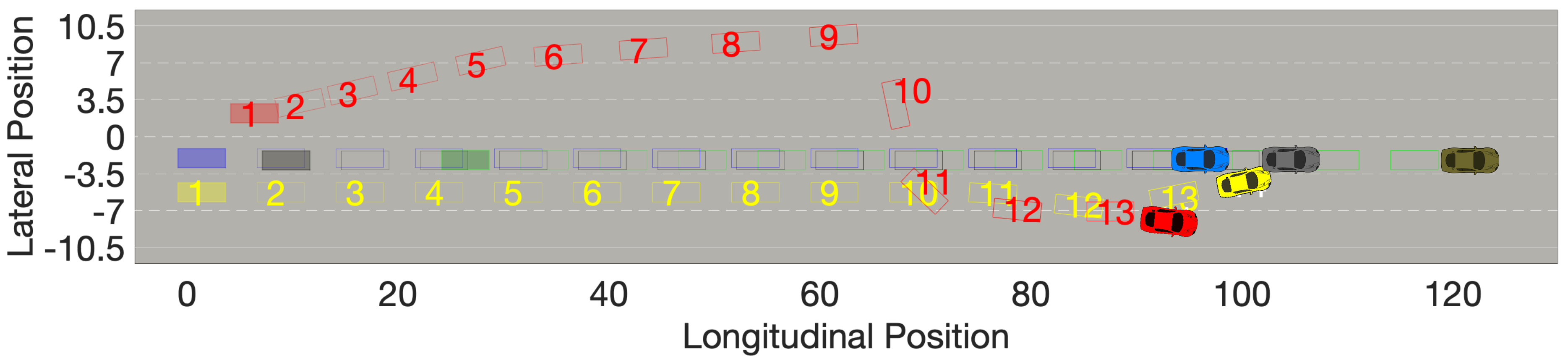}
		\caption{One of the collision cases discovered by the RRT-based approach.}
		\label{fig:rrt_collision_trace}
	\end{centering}
\end{figure}

\begin{figure}[htp]
	\begin{centering}
		\includegraphics[width=\columnwidth]{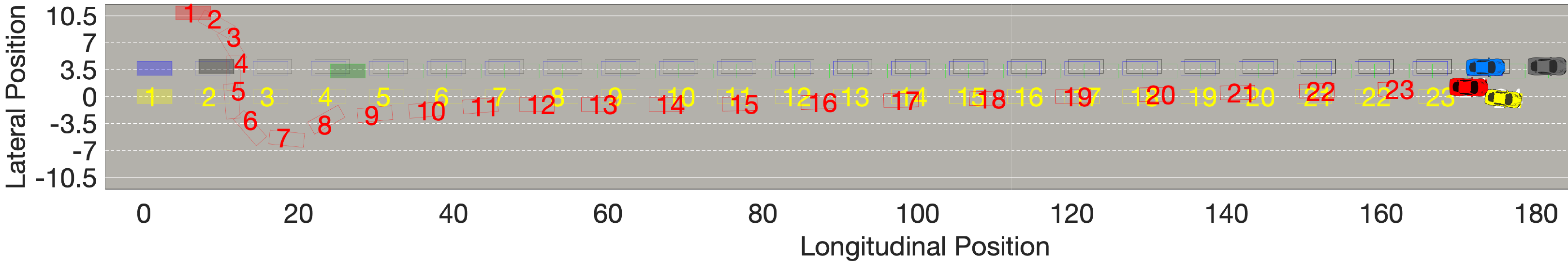}
		\caption{The small-speed collision discovered by the falsification-based approach.}
		\label{fig:fals_collision_trace}
	\end{centering}
\end{figure}

\begin{figure}[htp]
	\begin{centering}
		\includegraphics[width=0.4\columnwidth]{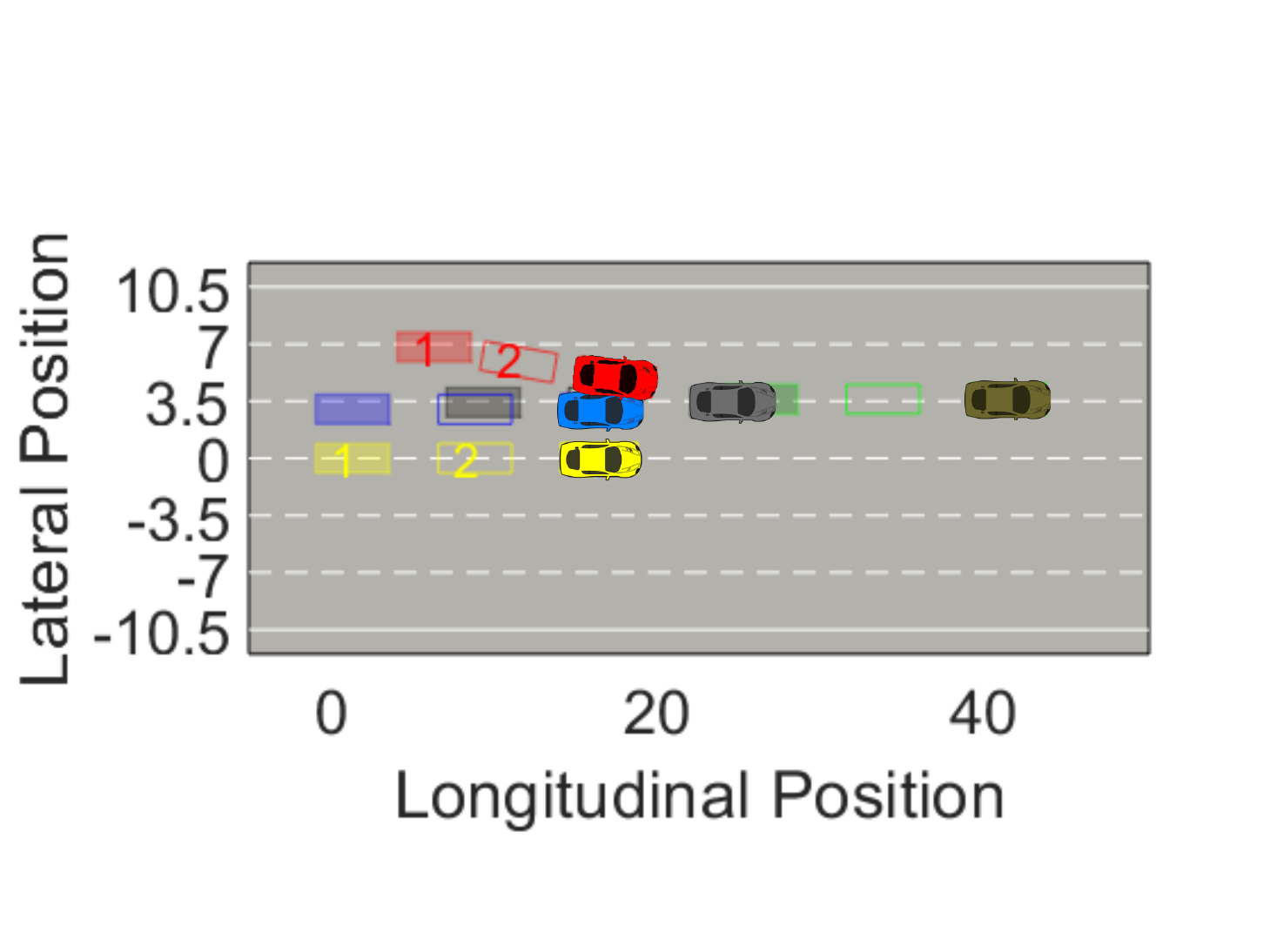}
		\caption{One of the cases where falsification-based approach was stuck in a local minimum.}
		\label{fig:fals_local_minima_trace}
	\end{centering}
\end{figure}

\figg{\ref{fig:rrt_collision_trace}} shows one of the interesting collision cases discovered by the RRT-based approach.
Agent $\agent_1$ first forces Ego to move to the right to avoid a collision and then to the left where it ends up colliding with Agent $\agent_3$.
Histories of $\agent_1$ (red) and Ego (yellow) vehicles are numbered to show their evolution over time.
\figg{\ref{fig:fals_collision_trace}} shows the only small-speed collision case discovered by the falsification-based approach.
Agent $\agent_1$ moves into the Ego vehicle's lane, accelerates and rear-ends with Ego vehicle even though Ego vehicle tries to avoid the collision by accelerating and steering away.
\figg{\ref{fig:fals_local_minima_trace}} shows a typical trajectory that is stuck in a local minimum.
Agent $\agent_1$ tries to move closer to Ego vehicle and reduces the time-to-collision but collides with one of the other agents, which is $\agent_2$ in this figure.
Although both approaches can get stuck in a local minimum, this case is significantly more common for the falsification-based approach as discussed above.

\begin{figure}[htp]
	\begin{centering}
		\includegraphics[width=\columnwidth]{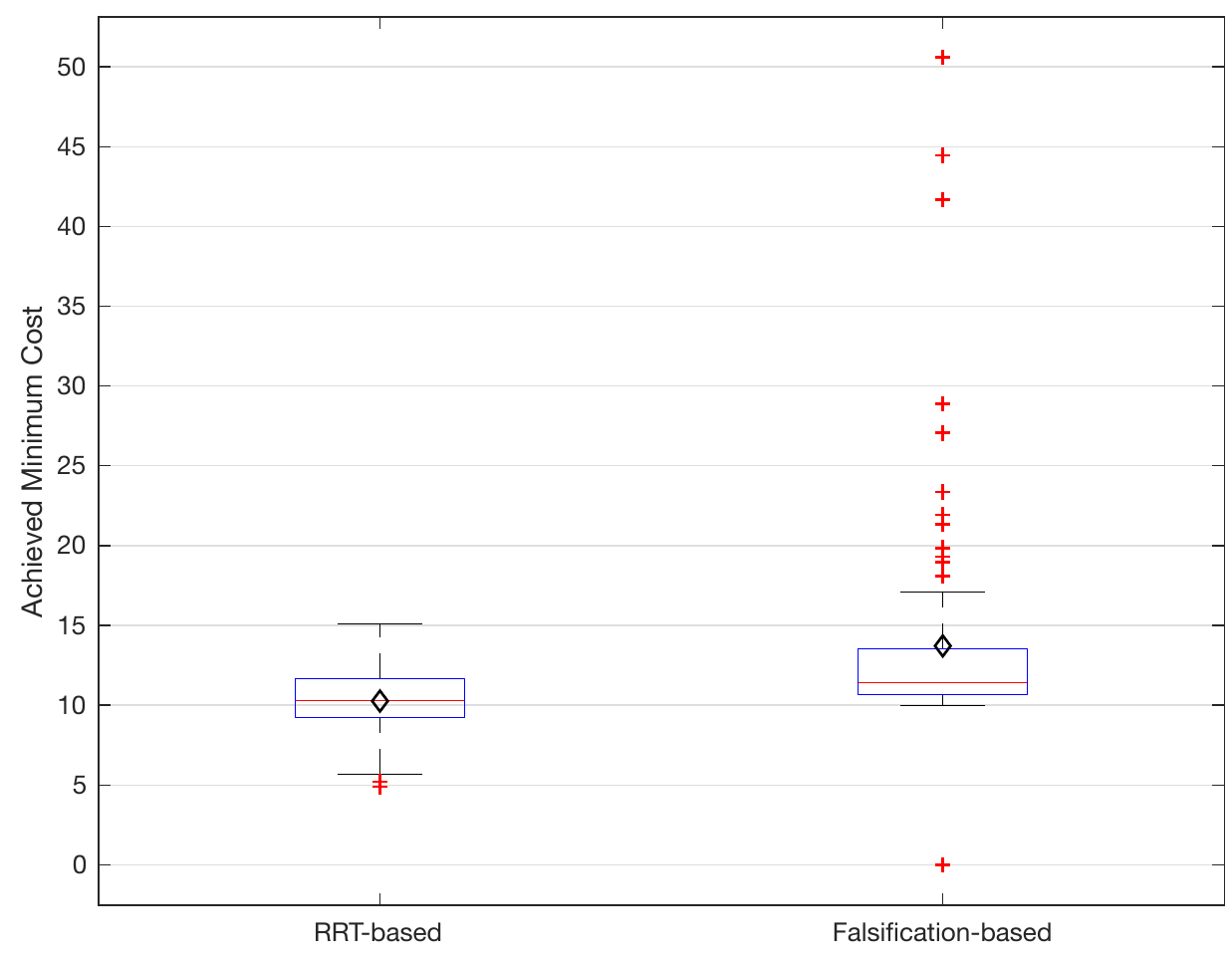}
		\caption{Comparison of the minimum cost achieved by the falsification approach and the RRT-based approach in Case Study 2.}
		\label{fig:boxplots}
	\end{centering}
\end{figure}

As a numerical comparison of the minimum costs discovered by the two approaches, the minimum, mean and maximum costs achieved by the falsification approach were $0.0043, 13.7134$, and $50.6017$, respectively.
The minimum, mean and maximum costs achieved by the RRT-based approach were $4.8955, 10.2571, 15.0856$.
\figg{\ref{fig:boxplots}} provides box and whisker diagrams of the minimum costs achieved by the two approaches among the $100$ experiments we have carried.
The black diamonds plotted on top of the box plots show the mean values for the returned minimum costs.
While $44$ out of $100$ RRT-based approach experiments were able to avoid local minima occurring around the cost $10$, only $2$ of the falsification approach experiments were able to achieve this.

An observation we would like to share is that in the single case where the falsification approach was able to cause a small speed collision, the achieved minimum cost was significantly smaller than any of the $11$ collision cases discovered by the RRT-based approach.
This observation supports our conclusion in Case Study 1, \ie when the falsification-based approach can get into the neighborhood of a local or global minimum, it is more able to get closer to the minimum point than the RRT-based approach.
The advantage of the RRT-based approach in avoiding local minima and the advantage of the falsification-based in reaching minima suggests that an approach combining these two has the potential to improve the overall test generation performance.
In such an approach, firstly, RRT-based approach would discover neighborhoods of minima and then the falsification-based approach would guide the test toward the minima starting from the results of the RRT-based approach.

%% file: RRT_approach_flowchart.tex
\begin{tikzpicture}[>=stealth',scale=1.0, every node/.style={scale=1.0}]

\tikzset{font={\fontsize{12pt}{12}\selectfont}}
\tikzstyle{stext}=[font=\fontsize{12}{12}\selectfont]
\tikzstyle{smalltext}=[font=\fontsize{12}{12}\selectfont]
\tikzstyle{block}=[draw,fill=white,rectangle,minimum height=2em,text
width=4.5em,fill=blue!10,stext,inner sep=4pt,align=center,execute at begin node=\setlength{\baselineskip}{1.2em}]
\tikzstyle{file}=[tape,fill=red!10,tape bend top=none,draw,inner
sep=2pt,text width=5em,align=center,smalltext,execute at begin node=\setlength{\baselineskip}{1.2em}]
\tikzstyle{note} = [coordinate]

\node[file, fill=green!20, text width=6em]  (init) {};

\node[block, below of=init, node distance=16mm, text width=16em] (describethescenario) {\textsc{Describe the Driving Scenario $(\setEgo, \setAgent, \setSurroundings, \stSet_0)$}};
\draw[->] (init) -- (describethescenario);

\node[block, below of=describethescenario, node distance=16mm, text width=16em] (sampleinitial) {\textsc{Sample Initial States}};
\draw[->] (describethescenario) -- (sampleinitial);

\node[block, below of=sampleinitial, node distance=14mm, text width=16em] (addconfig) {\textsc{Add current configuration\\ to the search tree $\rrtTree$}};
\draw[->] (sampleinitial) -- (addconfig);

\node[draw,smalltext,diamond,aspect=2,text width=6em,below of=addconfig,fill=yellow!10,node distance=18mm,inner sep=0pt,align=center] (decisiontermination) {Termination Condition?};
\draw[->] (addconfig) -- (decisiontermination);

\node[draw,rounded corners,align=center, right of=decisiontermination, node distance=45mm,fill=red!10,text width=4em] (halt) {\textsc{Halt}};
\draw[->] (decisiontermination.east) -- node[name=yes,above, near start]{Yes} (halt.west);

\node[block, below of=decisiontermination, node distance=18mm, text width=16em] (samplepath) {\textsc{Sample target path segments}};
\draw[->] (decisiontermination) -- node[name=decisionterminationNo, right]{No} (samplepath);

\node[block, below of=samplepath, node distance=14mm, text width=16em] (findbestnode) {\textsc{Find the best node in $\rrtTree$}};
\draw[->] (samplepath) -- (findbestnode);

\node[block, below of=findbestnode, node distance=14mm, text width=20em] (simulate) {\textsc{Simulate for $\searchdt$ starting from the best previous configuration}};
\draw[->] (findbestnode) -- (simulate);

\node[block, below of=simulate, node distance=16mm, text width=14em] (computecost) {\textsc{Compute Cost}};
\draw[->] (simulate) -- (computecost);

\node[draw,smalltext,diamond,aspect=2,text width=9em,below of=computecost,fill=yellow!10,node distance=25mm,inner sep=0pt,align=center] (decisiontransition) {\textsc{\ \\IsTransitionOK\\?}};
\draw[->] (computecost) -- node[name=newparam, right]{Cost} (decisiontransition);

\node[draw,smalltext,diamond,aspect=2,text width=5em,below of=decisiontransition,fill=yellow!10,node distance=30mm,inner sep=0pt,align=center] (decisionnovelty) {\textsc{\ \\IsNovel\\?}};
\draw[->] (decisiontransition) -- node[name=decisiontransitionYes, right]{Yes} (decisionnovelty);

\node[draw,smalltext,circle,aspect=2,text width=1em,left of=decisiontransition,fill=blue!10,node distance=50mm,inner sep=0pt,align=center] (junctionTransNovel) {};

\draw [->] (decisionnovelty.west) -| node[name=newparam, above right, near start, text width=6em,execute at begin node=\setlength{\baselineskip}{1.2em}]{\hspace*{9mm}No} (junctionTransNovel.south);

\draw [->] (decisiontransition.west) -- node[name=newparam, above right, near start, text width=3em,execute at begin node=\setlength{\baselineskip}{1.2em}]{No} (junctionTransNovel.east);

\draw [->] (junctionTransNovel.north) |- (decisiontermination.west) ;

\draw (decisionnovelty.south) -- node[name=newparam, right, text width=3em,execute at begin node=\setlength{\baselineskip}{1.2em}]{Yes} ($(decisionnovelty.south)+(0.0,-0.5)$) ;
\draw [->] ($(decisionnovelty.south)+(0.0,-0.5)$) -| ($(addconfig.west)+(-2.25,0.0)$) -- (addconfig.west) ;


\end{tikzpicture}

%% file: rrt_approach_conclusions.tex
\section{Conclusion and Future Work}
We proposed an approach that explores maneuvers for road occupants using rapidly-exploring random trees with the target of minimizing safety/performance cost functions defined for the automated driving system under test.
In our approach, we have adopted notions from transition-based RRT~\cite{jaillet2008transition} and RRT*~\cite{karaman2011sampling} methods.
Our RRT-based approach delivered more promising results compared to the stochastic optimization-based falsification approach which is described in \cite{tuncali2016itsc} for the problems which contain many local minima that should be avoided for reaching the global minimum.

When formulating the agent trajectory generation problem as an optimization problem, the trajectories are presented with a finite number of parameters, \ie waypoint parameters for a fixed number of waypoints, which are provided by the test designer.
One of the advantages of utilizing the RRT-based exploration is the ability to abandon the finite parameterization of the trajectories.
This creates the opportunity to more freely explore the space, and to minimize the need for manually designing the general structure of the trajectory shapes.
In an optimization-based approach, choosing a small number of parameters would limit the flexibility in generating critical trajectories, while choosing a large number would increase the dimensionality of the search space.
For instance, in our Case Study 2, the number of parameters that the optimization method would need to create the minimum-cost trajectories that are discovered by the RRT-based approach varies from $8$ to $68$ based on the number of parameters at each waypoint and the number of waypoints that are used to create these tajectories. 

Future work will include:
\begin{itemize}
	\item Exploring new methods for computing the novelty of a trajectory instead of a single state
	\item Computing a cost on the shape of the trajectory instead of computing the minimum of the instantaneous costs of the points on the trajectory.
	This may especially be useful for rewarding some types of vehicle paths such as the ones which are closer to real-world driving behaviors.
	\item Exploring new methods for using multi-objective optimization can be studied for different objectives like time-to-collision, collision speed, collision impact area etc.
	\item Combining RRT-based approach with the falsification-based approach such that the exploration starts with RRT-based approach and falsification-based approach further minimizes the cost starting from the best cases discovered by the RRT-based approach.
\end{itemize}

%% file: root.bbl
\begin{thebibliography}{10}
\providecommand{\bibitemdeclare}[2]{}
\providecommand{\surnamestart}{}
\providecommand{\surnameend}{}
\providecommand{\urlprefix}{Available at }
\providecommand{\url}[1]{\texttt{#1}}
\providecommand{\href}[2]{\texttt{#2}}
\providecommand{\urlalt}[2]{\href{#1}{#2}}
\providecommand{\doi}[1]{doi:\urlalt{http://dx.doi.org/#1}{#1}}
\providecommand{\bibinfo}[2]{#2}

\bibitemdeclare{article}{branicky2006sampling}
\bibitem{branicky2006sampling}
\bibinfo{author}{Michael~S \surnamestart Branicky\surnameend},
  \bibinfo{author}{Michael~M \surnamestart Curtiss\surnameend},
  \bibinfo{author}{Joshua \surnamestart Levine\surnameend} \&
  \bibinfo{author}{Stuart \surnamestart Morgan\surnameend}
  (\bibinfo{year}{2006}): \emph{\bibinfo{title}{Sampling-based planning,
  control and verification of hybrid systems}}.
\newblock {\sl \bibinfo{journal}{IEE Proceedings-Control Theory and
  Applications}} \bibinfo{volume}{153}(\bibinfo{number}{5}), pp.
  \bibinfo{pages}{575--590}.

\bibitemdeclare{book}{corke2017robotics}
\bibitem{corke2017robotics}
\bibinfo{author}{Peter \surnamestart Corke\surnameend} (\bibinfo{year}{2017}):
  \emph{\bibinfo{title}{Robotics, Vision and Control: Fundamental Algorithms In
  MATLAB{\textregistered} Second, Completely Revised}}.
\newblock \bibinfo{volume}{118}, \bibinfo{publisher}{Springer}.

\bibitemdeclare{inproceedings}{dang2008sensitive}
\bibitem{dang2008sensitive}
\bibinfo{author}{Thao \surnamestart Dang\surnameend},
  \bibinfo{author}{Alexandre \surnamestart Donz{\'e}\surnameend},
  \bibinfo{author}{Oded \surnamestart Maler\surnameend} \& \bibinfo{author}{Noa
  \surnamestart Shalev\surnameend} (\bibinfo{year}{2008}):
  \emph{\bibinfo{title}{Sensitive state-space exploration}}.
\newblock In: {\sl \bibinfo{booktitle}{Decision and Control, 2008. CDC 2008.
  47th IEEE Conference on}}, \bibinfo{organization}{IEEE}, pp.
  \bibinfo{pages}{4049--4054}.

\bibitemdeclare{article}{de2000mahalanobis}
\bibitem{de2000mahalanobis}
\bibinfo{author}{Roy \surnamestart De~Maesschalck\surnameend},
  \bibinfo{author}{Delphine \surnamestart Jouan-Rimbaud\surnameend} \&
  \bibinfo{author}{D{\'e}sir{\'e}~L \surnamestart Massart\surnameend}
  (\bibinfo{year}{2000}): \emph{\bibinfo{title}{The mahalanobis distance}}.
\newblock {\sl \bibinfo{journal}{Chemometrics and intelligent laboratory
  systems}} \bibinfo{volume}{50}(\bibinfo{number}{1}), pp.
  \bibinfo{pages}{1--18}.

\bibitemdeclare{inproceedings}{DreossiEtAl2017rmlw}
\bibitem{DreossiEtAl2017rmlw}
\bibinfo{author}{T.~\surnamestart Dreossi\surnameend},
  \bibinfo{author}{S.~\surnamestart Ghosh\surnameend},
  \bibinfo{author}{A.~\surnamestart Sangiovanni-Vincentelli\surnameend} \&
  \bibinfo{author}{S.~A. \surnamestart Seshia\surnameend}
  (\bibinfo{year}{2017}): \emph{\bibinfo{title}{Systematic Testing of
  Convolutional Neural Networks for Autonomous Driving}}.
\newblock In: {\sl \bibinfo{booktitle}{Reliable Machine Learning in the Wild
  (RMLW)}}.

\bibitemdeclare{inproceedings}{DreossiDDKJD15nfm}
\bibitem{DreossiDDKJD15nfm}
\bibinfo{author}{Tommaso \surnamestart Dreossi\surnameend},
  \bibinfo{author}{Thao \surnamestart Dang\surnameend},
  \bibinfo{author}{Alexandre \surnamestart Donz{\'{e}}\surnameend},
  \bibinfo{author}{James \surnamestart Kapinski\surnameend},
  \bibinfo{author}{Xiaoqing \surnamestart Jin\surnameend} \&
  \bibinfo{author}{Jyotirmoy~V. \surnamestart Deshmukh\surnameend}
  (\bibinfo{year}{2015}): \emph{\bibinfo{title}{Efficient Guiding Strategies
  for Testing of Temporal Properties of Hybrid Systems}}.
\newblock In: {\sl \bibinfo{booktitle}{7th International Symposium {NASA}
  Formal Methods {(NFM)}}}, {\sl \bibinfo{series}{LNCS}}
  \bibinfo{volume}{9058}, \bibinfo{publisher}{Springer}, pp.
  \bibinfo{pages}{127--142}.

\bibitemdeclare{incollection}{esposito2004adaptive}
\bibitem{esposito2004adaptive}
\bibinfo{author}{Joel~M \surnamestart Esposito\surnameend},
  \bibinfo{author}{Jongwoo \surnamestart Kim\surnameend} \&
  \bibinfo{author}{Vijay \surnamestart Kumar\surnameend}
  (\bibinfo{year}{2004}): \emph{\bibinfo{title}{Adaptive {RRTs} for validating
  hybrid robotic control systems}}.
\newblock In: {\sl \bibinfo{booktitle}{Algorithmic Foundations of Robotics
  VI}}, \bibinfo{publisher}{Springer}, pp. \bibinfo{pages}{107--121}.

\bibitemdeclare{inproceedings}{FainekosSUY12acc}
\bibitem{FainekosSUY12acc}
\bibinfo{author}{Georgios \surnamestart Fainekos\surnameend},
  \bibinfo{author}{Sriram \surnamestart Sankaranarayanan\surnameend},
  \bibinfo{author}{Koichi \surnamestart Ueda\surnameend} \&
  \bibinfo{author}{Hakan \surnamestart Yazarel\surnameend}
  (\bibinfo{year}{2012}): \emph{\bibinfo{title}{Verification of Automotive
  Control Applications using {S-TaLiRo}}}.
\newblock In: {\sl \bibinfo{booktitle}{Proceedings of the American Control
  Conference}}.

\bibitemdeclare{inproceedings}{hoffmann2007autonomous}
\bibitem{hoffmann2007autonomous}
\bibinfo{author}{Gabriel~M \surnamestart Hoffmann\surnameend},
  \bibinfo{author}{Claire~J \surnamestart Tomlin\surnameend},
  \bibinfo{author}{Michael \surnamestart Montemerlo\surnameend} \&
  \bibinfo{author}{Sebastian \surnamestart Thrun\surnameend}
  (\bibinfo{year}{2007}): \emph{\bibinfo{title}{Autonomous automobile
  trajectory tracking for off-road driving: Controller design, experimental
  validation and racing}}.
\newblock In: {\sl \bibinfo{booktitle}{American Control Conference}}, pp.
  \bibinfo{pages}{2296--2301}.

\bibitemdeclare{inproceedings}{jaillet2008transition}
\bibitem{jaillet2008transition}
\bibinfo{author}{L{\'e}onard \surnamestart Jaillet\surnameend},
  \bibinfo{author}{Juan \surnamestart Cort{\'e}s\surnameend} \&
  \bibinfo{author}{Thierry \surnamestart Sim{\'e}on\surnameend}
  (\bibinfo{year}{2008}): \emph{\bibinfo{title}{Transition-based {RRT} for path
  planning in continuous cost spaces}}.
\newblock In: {\sl \bibinfo{booktitle}{Intelligent Robots and Systems, 2008.
  IROS 2008. IEEE/RSJ International Conference on}},
  \bibinfo{organization}{IEEE}, pp. \bibinfo{pages}{2145--2150}.

\bibitemdeclare{article}{karaman2011sampling}
\bibitem{karaman2011sampling}
\bibinfo{author}{Sertac \surnamestart Karaman\surnameend} \&
  \bibinfo{author}{Emilio \surnamestart Frazzoli\surnameend}
  (\bibinfo{year}{2011}): \emph{\bibinfo{title}{Sampling-based algorithms for
  optimal motion planning}}.
\newblock {\sl \bibinfo{journal}{The international journal of robotics
  research}} \bibinfo{volume}{30}(\bibinfo{number}{7}), pp.
  \bibinfo{pages}{846--894}.

\bibitemdeclare{techreport}{kim2005rrt}
\bibitem{kim2005rrt}
\bibinfo{author}{Jongwoo \surnamestart Kim\surnameend}, \bibinfo{author}{Joel~M
  \surnamestart Esposito\surnameend} \& \bibinfo{author}{Vijay \surnamestart
  Kumar\surnameend} (\bibinfo{year}{2005}): \emph{\bibinfo{title}{An
  {RRT}-based algorithm for testing and validating multi-robot controllers}}.
\newblock \bibinfo{type}{Technical Report}, \bibinfo{institution}{Moore Schoolf
  Of Electrical Engineering Philadelphi PA Grasp Lab}.

\bibitemdeclare{techreport}{lavalle1998rapidly}
\bibitem{lavalle1998rapidly}
\bibinfo{author}{Steven~M \surnamestart LaValle\surnameend}
  (\bibinfo{year}{1998}): \emph{\bibinfo{title}{Rapidly-exploring random trees:
  A new tool for path planning}}.
\newblock \bibinfo{type}{Technical Report}.

\bibitemdeclare{inproceedings}{lehman2008exploiting}
\bibitem{lehman2008exploiting}
\bibinfo{author}{Joel \surnamestart Lehman\surnameend} \&
  \bibinfo{author}{Kenneth~O \surnamestart Stanley\surnameend}
  (\bibinfo{year}{2008}): \emph{\bibinfo{title}{Exploiting open-endedness to
  solve problems through the search for novelty.}}
\newblock In: {\sl \bibinfo{booktitle}{ALIFE}}, pp. \bibinfo{pages}{329--336}.

\bibitemdeclare{article}{plaku2009hybrid}
\bibitem{plaku2009hybrid}
\bibinfo{author}{Erion \surnamestart Plaku\surnameend},
  \bibinfo{author}{Lydia~E \surnamestart Kavraki\surnameend} \&
  \bibinfo{author}{Moshe~Y \surnamestart Vardi\surnameend}
  (\bibinfo{year}{2009}): \emph{\bibinfo{title}{Hybrid systems: from
  verification to falsification by combining motion planning and discrete
  search}}.
\newblock {\sl \bibinfo{journal}{Formal Methods in System Design}}
  \bibinfo{volume}{34}(\bibinfo{number}{2}), pp. \bibinfo{pages}{157--182}.

\bibitemdeclare{article}{stanley2002evolving}
\bibitem{stanley2002evolving}
\bibinfo{author}{Kenneth~O \surnamestart Stanley\surnameend} \&
  \bibinfo{author}{Risto \surnamestart Miikkulainen\surnameend}
  (\bibinfo{year}{2002}): \emph{\bibinfo{title}{Evolving neural networks
  through augmenting topologies}}.
\newblock {\sl \bibinfo{journal}{Evolutionary computation}}
  \bibinfo{volume}{10}(\bibinfo{number}{2}), pp. \bibinfo{pages}{99--127}.

\bibitemdeclare{phdthesis}{tuncali2019dissertation}
\bibitem{tuncali2019dissertation}
\bibinfo{author}{Cumhur~Erkan \surnamestart Tuncali\surnameend}
  (\bibinfo{year}{2019}): \emph{\bibinfo{title}{Search-based Test Generation
  for Automated Driving Systems: From Perception to Control Logic}}.
\newblock Ph.D. thesis, \bibinfo{school}{Arizona State University}.

\bibitemdeclare{inproceedings}{tuncali2018iv}
\bibitem{tuncali2018iv}
\bibinfo{author}{Cumhur~Erkan \surnamestart Tuncali\surnameend},
  \bibinfo{author}{Georgios \surnamestart Fainekos\surnameend},
  \bibinfo{author}{Hisahiro \surnamestart Ito\surnameend} \&
  \bibinfo{author}{James \surnamestart Kapinski\surnameend}
  (\bibinfo{year}{2018}): \emph{\bibinfo{title}{Simulation-based Adversarial
  Test Generation for Autonomous Vehicles with Machine Learning Components}}.
\newblock In: {\sl \bibinfo{booktitle}{{IEEE} Intelligent Vehicles Symposium
  ({IV})}}.

\bibitemdeclare{inproceedings}{tuncali2016itsc}
\bibitem{tuncali2016itsc}
\bibinfo{author}{Cumhur~Erkan \surnamestart Tuncali\surnameend},
  \bibinfo{author}{Theodore~P. \surnamestart Pavlic\surnameend} \&
  \bibinfo{author}{Georgios \surnamestart Fainekos\surnameend}
  (\bibinfo{year}{2016}): \emph{\bibinfo{title}{Utilizing {S-TaLiRo} as an
  Automatic Test Generation Framework for Autonomous Vehicles}}.
\newblock In: {\sl \bibinfo{booktitle}{IEEE Intelligent Transportation Systems
  Conference}}.

\end{thebibliography}
